\documentclass[]{sensenova}

\usepackage{hyperref}
\usepackage{cleveref}
\usepackage{verbatim}

\usepackage{wrapfig}
\usepackage{graphicx}
\usepackage{floatrow}
\usepackage{placeins}
\usepackage{subcaption}
\usepackage{listings}
\usepackage{algorithm}
\usepackage{subcaption}
\usepackage{dsfont}

\usepackage{mmstyles}

\usepackage[toc,page,header]{appendix}

\usepackage{xspace}
\usepackage[utf8]{inputenc}  
\usepackage[T1]{fontenc}     
\usepackage{CJKutf8} 

\usepackage[misc]{ifsym}

\usepackage{CJKutf8} 
\usepackage[colorinlistoftodos,prependcaption,textsize=small]{todonotes}
\newcommand{\MO}[1]{\textcolor{black}{#1}}
\usepackage{fontawesome5}

\usepackage{soul} 
\definecolor{HLGreen}{rgb}{0.78,0.95,0.78}
\definecolor{HLRed}{rgb}{0.9725,0.8431,0.8549}   
\definecolor{benchblue}{RGB}{58, 122, 252}

\newcommand{\colornum}[1]{%
  \footnotesize
  \ifdim #1 pt < 0pt
    \textcolor{sensepurple}{#1}%
  \else
    \textcolor{benchblue}{#1}%
  \fi
}

\usepackage{cuted} 

\definecolor{bestcolor}{RGB}{219, 208, 237}
\definecolor{secondcolor}{RGB}{241, 237, 248}
\definecolor{thirdcolor}{RGB}{211, 222, 190}
\definecolor{line-blue}{RGB}{243, 248, 252}
\definecolor{line-green}{RGB}{200,242,200}
\definecolor{line-red}{RGB}{255,215,215}

\usepackage{adjustbox}
\usepackage{makecell}
\usepackage{booktabs}

\newlength{\ModelW}
\newlength{\NonModelW}
\newcolumntype{M}{>{\centering\arraybackslash}p{\NonModelW}}
\usepackage{graphicx}
\usepackage{subcaption}
\graphicspath{{assets/figures/}}

\usepackage{array}      
\usepackage{caption}
\usepackage{subcaption} 

\usepackage{enumitem}

\usepackage{multirow}
\usepackage{makecell}

\usepackage{booktabs,tabularx,array}
\newcolumntype{Y}{>{\centering\arraybackslash}X}
\newcolumntype{L}{>{\raggedright\arraybackslash}X}

\usepackage{framed}
\usepackage{algorithm}
\usepackage{algorithmic}
\usepackage{listings}
\usepackage{soul}          
\usepackage{wrapfig}
\usepackage{stfloats}

\definecolor{cotredbg}{RGB}{255,220,220}   
\definecolor{cotgreenbg}{RGB}{220,255,220} 

\title{Demystifying Video Reasoning} 

\author{
Ruisi Wang$^{1}$, Zhongang Cai$^{\textrm{\Letter},1}$, Fanyi Pu$^{1,2}$, Junxiang Xu$^{1}$, Wanqi Yin$^{1}$, \\Maijunxian Wang$^{3}$, Ran Ji$^{4}$, Chenyang Gu$^{1}$, Bo Li$^{2}$, Ziqi Huang$^{2}$, \\Hokin Deng$^{5}$, Dahua Lin$^{1}$, Ziwei Liu$^{2}$, Lei Yang$^{1}$

\parbox{\textwidth}{\centering\small
    $\textrm{\Letter}$ Corresponding Author \\
    $^{1}$ SenseTime Research \quad
    $^{2}$ Nanyang Technological University \quad 
    $^{3}$ University of California, Berkeley \quad \\
    $^{4}$ University of California, San Diego \quad
    $^{5}$ Carnegie Mellon University}}
\abstract{

Recent advances in video generation have revealed an unexpected phenomenon: diffusion-based video models exhibit non-trivial reasoning capabilities. Prior work attributes this to a Chain-of-Frames (CoF) mechanism, where reasoning is assumed to unfold sequentially across video frames. In this work, we challenge this assumption and uncover a fundamentally different mechanism. We show that reasoning in video models instead primarily emerges along the \textit{diffusion denoising steps}. Through qualitative analysis and targeted probing experiments, we find that models explore multiple candidate solutions in early denoising steps and progressively converge to a final answer, a process we term \textbf{Chain-of-Steps (CoS)}.
Beyond this core mechanism, we identify several emergent reasoning behaviors critical to model performance: (1) \textbf{working memory} that supports tasks requiring \MO{consistent} reference, such as object permanence; (2) \textbf{self-correction and enhancement}, allowing recovery from incorrect intermediate solutions; and (3) \textbf{perception before action}, where early steps establish semantic grounding and later steps perform structured manipulation. 
\MO{Moreover, analysis of Diffusion Transformer layers shows that middle layers conduct key reasoning procedures.}
Motivated by these insights, we present a simple \textbf{Training-Free Ensemble (TFE)} as a proof-of-concept, demonstrating how reasoning can be improved by ensembling latent trajectories from identical models with different random seeds.
Overall, our work provides the first systematic dissection of the mechanisms underlying video reasoning, offering a foundation to guide future research in better exploiting the inherent reasoning dynamics of video models as a new substrate for intelligence.
\vspace{\baselineskip}

\textbf{keywords: }Video Reasoning · Diffusion Models · Emergent Intelligence

\checkdata[Homepage]{\url{https://www.wruisi.com/demystifying_video_reasoning}}
}

\begin{document}
\maketitle

\begin{figure}
\centering
\vspace{-6mm}
\includegraphics[width=\linewidth]{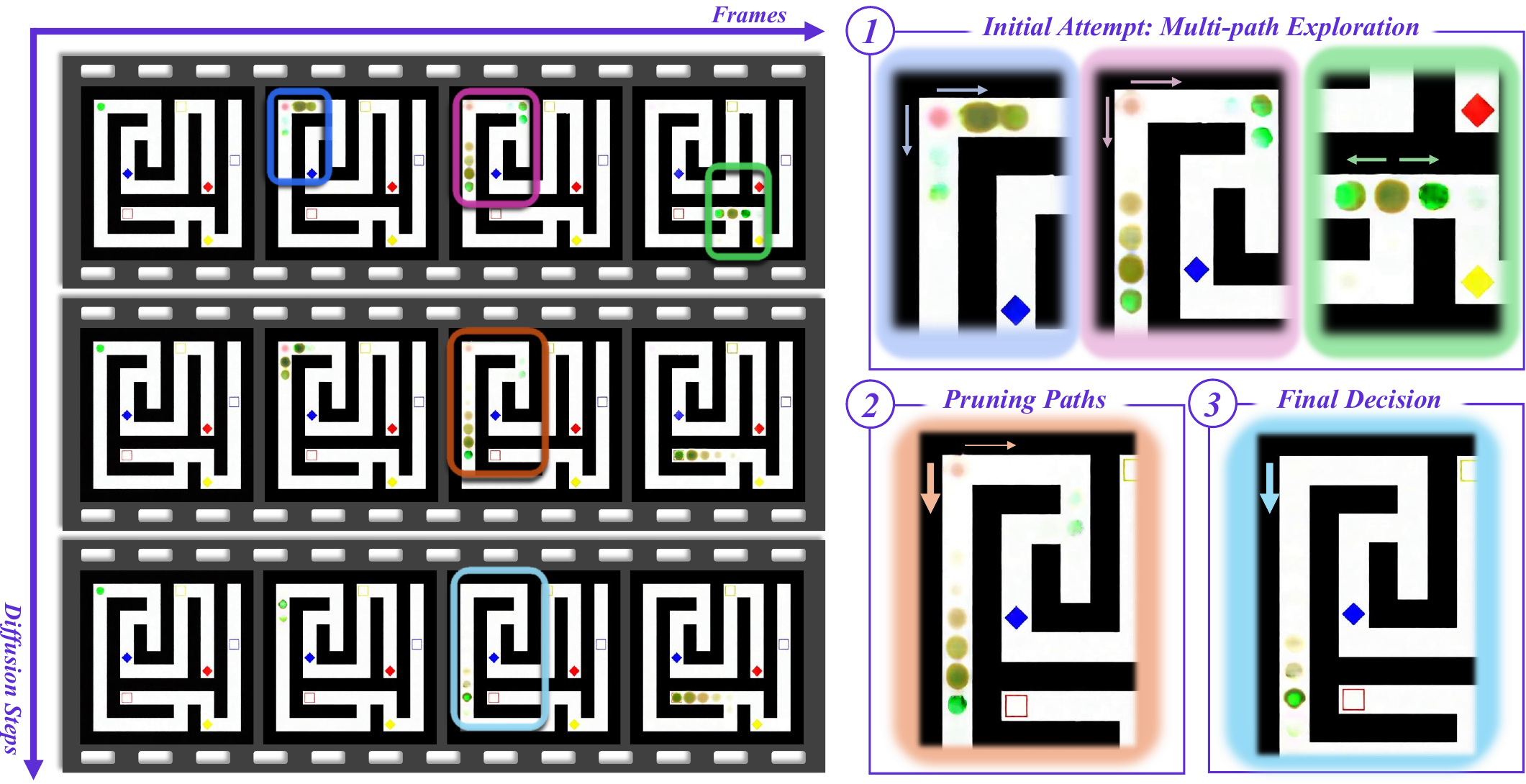}
\vspace{-4mm}
\caption{
\textbf{Chain-of-Steps.}
We discover that video reasoning occurs along the diffusion steps with surprising emergent behaviors such as making multiple possible moves (\eg, paths) simultaneously at early steps, gradually pruning suboptimal choices during middle steps, and reaching a final decision at the late steps. This maze-solving example asks the model to start from the green circle in the top-left corner and find the red rectangle. Key regions of interest are color-coded and enlarged on the right. 
}
\label{fig:teaser}

\end{figure}

\section{Introduction}
\label{sec:intro}

Video generation models have transformed the landscape of movie, gaming, and entertainment industries. However, most research has focused primarily on their ability to produce high-fidelity, realistic, and visually appealing videos. Recent advances have revealed an unexpected phenomenon: diffusion-based video models exhibit non-trivial reasoning capabilities in spatiotemporally consistent visual environments~\cite{wiedemer2025video}. Prior work attributes this behavior to a Chain-of-Frames (CoF) mechanism, suggesting that reasoning unfolds sequentially across video frames. Despite this intriguing discovery, the underlying mechanisms of video reasoning remain largely unexplored.
With the recent release of large-scale video reasoning datasets and open-source foundation models~\cite{vbvr2026}, we can now systematically investigate this capability. Leveraging these resources, we conduct the first comprehensive dissection of video reasoning and uncover a fundamentally different mechanism: \textbf{reasoning in diffusion-based video models primarily emerges along the denoising process} rather than across frames. 

%
Our key discovery challenges the prevailing Chain-of-Frames (CoF) hypothesis~\cite{wiedemer2025video, wu2025chronoedit}, which assumes that video reasoning unfolds sequentially across frames. Instead, we find that reasoning does not primarily operate along the temporal dimension. Rather, it emerges along the diffusion denoising steps, progressing throughout generation. We term this mechanism \textbf{Chain-of-Steps (CoS)}.
This finding suggests a fundamentally different view of how diffusion-based video models reason. Due to bidirectional attention over the entire sequence, reasoning is performed across all frames simultaneously at each denoising step, with intermediate hypotheses progressively refined as the process unfolds.
Qualitative analysis reveals intriguing dynamics. In early denoising steps, the model often entertains multiple possibilities, exploring alternative trajectories before progressively committing to one. Importantly, we view reasoning not as the mere coexistence of multiple possible outcomes, but as a directed process of selecting and refining a solution from among candidates.
Moreover, noise perturbation analysis shows that disruptions at specific denoising steps significantly degrade performance, whereas frame-wise perturbations have a much weaker impact. Further information propagation analysis identifies that conclusions primarily solidify during the middle diffusion steps.

Furthermore, we uncover several surprising emergent behaviors in video reasoning models that are strikingly similar to those observed in early studies of Large Language Models (LLMs). First, these models exhibit a form of \textbf{working memory} that is crucial for tasks requiring persistent references (\eg, object permanence). Second, we observe that video models can \textbf{self-correct} errors during the CoS reasoning process, rather than committing to incorrect trajectories. Third, video models exhibit a \textbf{"perception before action"} behavior, where early diffusion steps prioritize localizing target objects before subsequent steps perform more complex reasoning and manipulation.

We further conduct a fine-grained analysis of the Diffusion Transformer by examining token representations within a single diffusion step. This reveals self-evolved,  diverse, task-agnostic functional layers throughout the network. Within one step, early layers focus on dense perceptual understanding (\eg, separating foreground from background and identifying basic geometric structures), while a set of critical middle layers performs most of the reasoning. The final layers then consolidate the latent representation to produce the next step video state.

Motivated by these insights, we present \MO{Training-Free Ensemble (TFE)}, a simple method as a proof-of-concept for improving video reasoning models. Given that the model inherently explores multiple reasoning paths during the diffusion process, we propose an inference-time ensemble strategy that merges latents produced by three identical models with different random seeds. This encourages the preservation of a richer set of candidate reasoning trajectories during generation. As a result, the model explores more diverse reasoning paths and is more likely to converge to the correct solution, 
illustrating \MO{how our findings can inform more effective video reasoning systems}. 

In summary, we investigate the underlying mechanisms of video reasoning in diffusion models and identify Chain-of-Steps (CoS), a reasoning process that unfolds along the denoising trajectory. We further uncover several emergent reasoning behaviors that arise in these models. Building on these insights, we demonstrate how such mechanisms can be exploited through a simple training-free strategy for reasoning path ensembling. We believe our findings provide a foundation for understanding and advancing video reasoning, positioning it as a promising next-generation substrate for machine intelligence.

\section{Related Works}
\label{sec:related_works}

\subsection{Reasoning in Language and Multimodal Models}

Recent studies show that large language models (LLMs) exhibit remarkable reasoning capabilities. Early work identifies emergent behaviors that arise as models scale in size and data~\cite{wei2022emergent}, and demonstrates that Chain-of-Thought (CoT) prompting, which elicits intermediate reasoning steps, significantly improves performance~\cite{wei2022chain}. Subsequent work explores mechanisms such as self-reflection, correction, and action~\cite{madaan2023selfrefine, yao2023react, huang2023large, yang2025understanding}. Coconut further suggests that reasoning can also occur implicitly within latent representations~\cite{coconut}.
Meanwhile, research has increasingly explored extending reasoning beyond language into multimodal settings. Early progress in vision-language models (VLMs) enables reasoning over images in addition to text~\cite{flamingo, blip2, llava, qwenvl, internvl}, whereas recent work has studied unified architectures that jointly model language and vision~\cite{seedx, chameleon, tong2025metamorph, qu2025tokenflow, bagel, wu2025omnigen2, wu2024janus, zhuang2025vargpt, chen2025blip3o, metaqueries, li2025unifork, zou2025uni}. These architectures empower reasoning for generation~\cite{koh2023generating, shi2024lmfusion, xiao2025mindomni, duan2025got, guo2025can, liao2025mogao, zheng2023minigpt5}, enable reasoning with generation through visual CoT~\cite{mvot, fang2025got, chern2025thinking, shi2025mathcanvas, t2i-r1, xu2025visualplanningletsthink, qin2025unicot, vchain_huang2025vchain, chen2026unit}, and extend to embodied scenarios~\cite{zhao2025cot, Zawalski24-ecot, zeng2025futuresightdrive}.
Together, these findings suggest that reasoning over multimodal signals opens up avenues for advanced reasoning capabilities. However, these efforts remain limited to discrete text and static images, making it challenging to leverage spatiotemporally consistent priors. Our work aims to investigate video as the next substrate for reasoning in intelligent systems.

\subsection{Video Generation Models}

Video generation has advanced rapidly with the development of diffusion models~\cite{ddpm_NEURIPS2020_4c5bcfec, latentdiffusion_rombach2022high} and high-fidelity variational autoencoders (VAEs)~\cite{vae_kingma2013auto, vaevideogen_denton2018stochastic, cvvae_zhao2024cv}. While early approaches focus primarily on generating short clips, the emergence of Diffusion Transformers (DiTs)~\cite{dit_peebles2023scalable} has enabled effective scaling of data and model size. As a result, recent video generators~\cite{kong2024hunyuan, wan_wan2025wan, veo3.1_deepmind_2026, sora_openai_2025, kling_ai_kuai_2025, runway_gen4_2025, lavie_wang2025lavie, fan2025vchitect} achieve impressive visual fidelity.
Despite these advances, major challenges remain in physical plausibility~\cite{yue2025simulating, zheng2025vbench2}, commonsense knowledge~\cite{zheng2025vbench2}, and spatiotemporal reasoning~\cite{wiedemer2025video, vreasonbench_luo2025v, vbvr2026}. Consequently, recent research has begun shifting toward investigating the reasoning capabilities of video generation models.
One line of work leverages the reasoning abilities of multimodal LLMs to guide video synthesis. For example, VChain~\cite{vchain_huang2025vchain} and MetaCanvas~\cite{metacanvas_lin2025exploring} incorporate external reasoning modules into pre-trained generators, while Omni-Video~\cite{omnivideo_tan2025omni} uses symbolic reasoning from LLMs to guide generation.
More recently, several studies ask whether video generators themselves can perform reasoning without external supervision, treating them as zero-shot learners operating in spatiotemporal environments~\cite{wiedemer2025video, thinkwithvideo_tong2025thinking, rulerbench_he2025ruler}. However, the mechanisms underlying this capability remain unexplored. Our work addresses this gap by investigating the internal reasoning processes of diffusion-based video models.

\subsection{Similarities to Biological Brains}
The diffusion model may exhibit behavior analogous to planning processes in biological brains. For instance, when a rat is deciding how to navigate toward a food reward, studies have shown that the hippocampus sequentially replays multiple candidate trajectories during a deliberation period before movement begins \cite{pfeiffer2013hippocampal}. Recent work suggests that human cognition may rely on analogous mechanisms of internal simulation and prospective planning to support conceptual reasoning, inference, and decision-making \cite{behrens2018cognitive, mattar2022planning}.
\section{Chain-of-Steps: Reasoning along Diffusion Steps}
\label{sec:dimension_of_reasoing}
\begin{figure*}[t!]
    \centering
    \includegraphics[width=\textwidth]{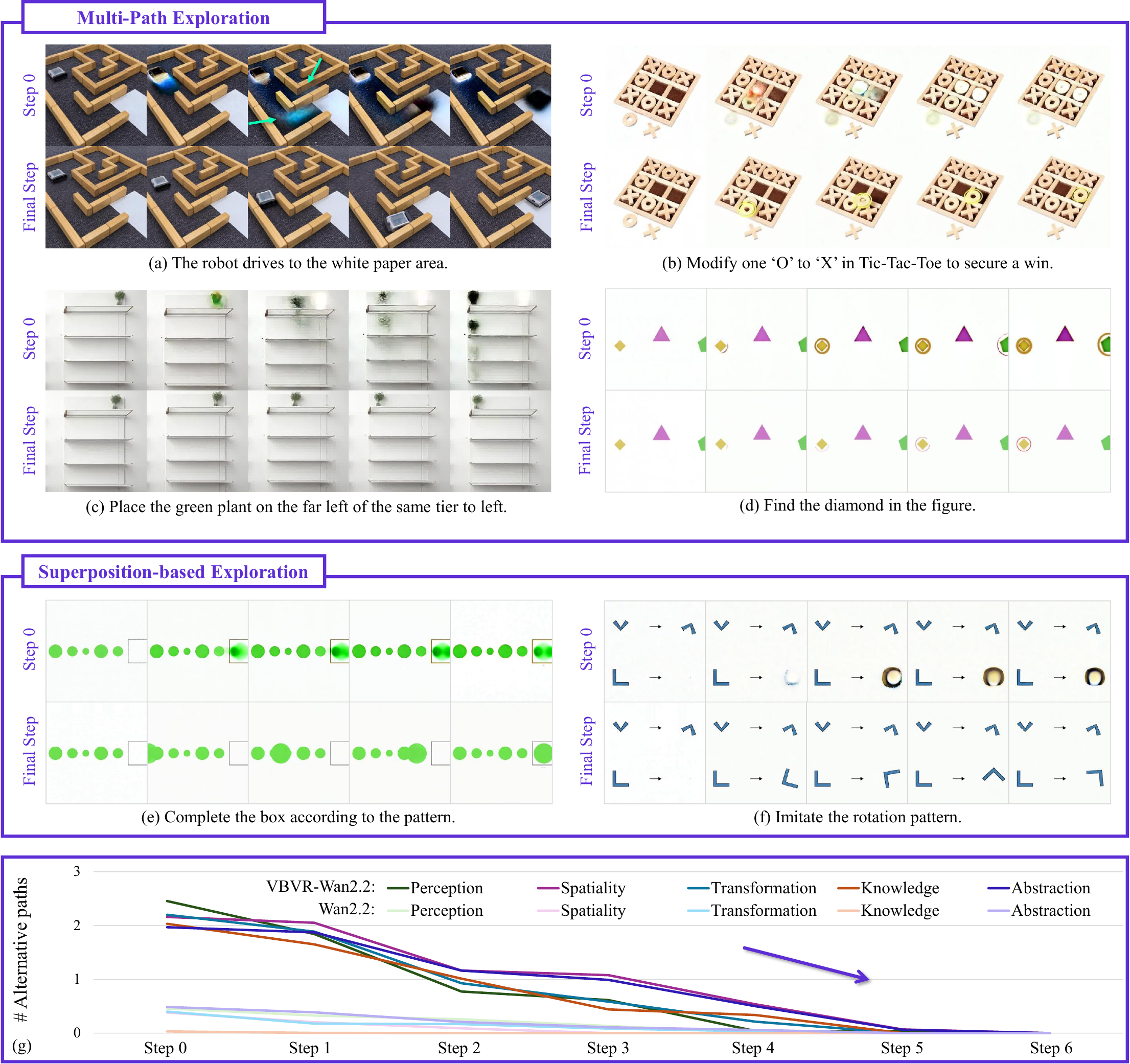}
    \setlength{\abovecaptionskip}{-6pt}
    \caption{ 
    \textbf{Chain-of-Steps elicits reasoning along the diffusion process.} We observe that video reasoning models explore multiple possible solutions simultaneously in the early denoising steps before converging to a final outcome in later steps. Specifically, we observe: (a) two potential routes (cyan arrows highlight the "imaginary traces") for the robot; (b) two possible placements of the "O" piece; (c) multiple candidate end positions for the plant; (d) simultaneous selection of two diamonds; (e) large and small circles overlapping with each other; and (f) all possible rotations of the L-shaped object superimposed. (g) Average number of alternative intermediate candidates observed at each diffusion step.}
    \setlength{\belowcaptionskip}{-6pt}
    \label{fig:multi_path}
\end{figure*}
While prior work~\cite{wiedemer2025video} hypothesizes a \textit{Chain-of-Frames} (CoF) mechanism in which reasoning in video models unfolds frame by frame, generated frames appear to exhibit a “causal” property where later frames gradually build conclusions conditioned on earlier frames. However, our analysis of the underlying video reasoning mechanism reveals evidence to the contrary.
First, we empirically analyze a wide range of reasoning tasks and find that the core logical reasoning in video generation models occurs across the diffusion denoising steps (\cref{sec:cos}). Diffusion steps do more than merely refine visual texture; instead, they explore multiple possibilities, evaluate their plausibility, and gradually converge to the correct outcome through the denoising process.
Second, we introduce noise perturbations to disrupt information flow at both the frame and step levels (\cref{sec:noise_perturbation}). Our findings reaffirm that CoS, rather than CoF, more accurately characterizes the reasoning mechanism in video models.
\MO{To ensure generality, we analyze multiple base models (\ie, LTX2.3~\cite{hacohen2026ltx2}, Wan2.1~\cite{wan_wan2025wan}, and Wan2.2~\cite{wan_wan2025wan}) and their reasoning-enhanced variants (with LoRA tuned on VBVR~\cite{vbvr2026}, indicated by the \textbf{VBVR-} prefix) that are significantly better at instruction-following and precise manipulation.
%
%
Unless otherwise stated, visualizations shown in the main paper are based on VBVR-Wan2.2, which is the strongest video reasoning model based on Wan2.2-I2V-A14B~\cite{wan_wan2025wan}, while results for more models are provided in \cref{sec:ltx2.3_and_wan2.1}.}

\subsection{Diffusion Steps as the Primary Axis of Reasoning}
\label{sec:cos}

\MO{To understand how semantic decisions emerge during denoising, we seek to visualize the model's intermediate trajectories throughout the diffusion process. However, this is challenging because diffusion models do not explicitly expose interpretable reasoning states. Directly decoding the intermediate latent $x_s$ is noisy and uninformative, as flow matching~\cite{lipman2022flow} constructs $x_s$ as a linear interpolation between semantic content and stochastic noise,}
\begin{equation}
  x_s = (1 - s)x_0 + sx_1
\end{equation}
\MO{where $x_0$ is the clean latent and $x_1 \sim \mathcal{N}(\mathbf{0}, \mathbf{I})$ is noise. To obtain a more interpretable view, we instead compute the estimated clean latent, using the predicted velocity field $v_\theta(x_s, s, c)$ and the corresponding noise scale $\sigma_s$.}
\begin{equation}
  \hat{x}_0 = x_s - \sigma_s \cdot v_\theta(x_s, s, c)
\end{equation}
\MO{Importantly, $\hat{x}_0$ is not a latent encountered during inference, but rather the model's instantaneous prediction of the final denoised sample given all information available at diffusion step s. Decoding $\hat{x}_0$ therefore provides an interpretable snapshot of the model's current hypothesis, enabling us to trace how semantic decisions evolve and analyze the model’s intermediate reasoning dynamics. }

\MO{Visualization examples are drawn from video reasoning benchmarks such as VBVR~\cite{vbvr2026} and general video generation benchmarks such as VBench~\cite{huang2023vbench,vbench_huang2025vbench++}.} Analogous to LLMs that exhibit reasoning behaviors along chain-of-thought, where the model gradually reaches its conclusion, to our surprise, we discover a similar scheme in video reasoning models along diffusion denoising steps.
This is exemplified in \cref{fig:teaser}, for complex navigational tasks such as maze-solving, early decoded predictions $\hat{x}_0$ appear as a probabilistic cloud in which several plausible paths are spawned and explored in parallel. Over subsequent steps, suboptimal trajectories gradually get suppressed, converging towards the final solution. \MO{By analyzing intermediate predictions at each step, we move beyond the temporal "Chain-of-Frames" (CoF) perspective and identify two distinct modes of Step-wise Reasoning: Multi-path Exploration and Superposition-based Exploration.}

\subsubsection{Multi-Path Exploration.}
\label{sec:multi_path}

In high-complexity logical tasks, the diffusion process resembles a Breadth-First Search (BFS) or a multi-choice elimination procedure, where the model explores a tree of possible solutions and gradually prunes incorrect branches.
It is worth noting that this behavior is reminiscent of the parallel reasoning trajectories explicitly studied in the LLM community (\eg, Tree of Thoughts~\cite{NEURIPS2023_271db992}). However, video generation models naturally explore multiple solution paths in parallel during the diffusion process, inherently performing a similar form of structured search within their latent space.
In some tasks involving object movements, the model explicitly visualizes this exploration process through multiple motion trajectories. In other tasks where the model must select an action from a discrete set of alternatives, we observe that the model initially considers several actions simultaneously and progressively discards candidates as the denoising process proceeds, until only a single valid outcome remains. 



\begin{itemize}
    \item \textit{\cref{fig:multi_path}(a) Robot Navigation.} The intermediate steps show the robot simultaneously exploring both the upper and lower routes through the maze. As the diffusion process proceeds, the trajectory corresponding to the lower path becomes increasingly dominant, while the alternative route gradually disappears, indicating that the model chooses the final path.
    \item \textit{\cref{fig:multi_path}(b) Tic-Tac-Toe.} During the early reasoning stage, the model simultaneously highlights multiple candidate cells for a winning move.
    \item \textit{\cref{fig:multi_path}(c) Object Movement.} In this example, it is clearly observable that at the early stage, the model proposes four potential trajectories corresponding to the four layers on the left side of the shelf. As the denoising steps continue, these alternatives gradually collapse toward placing the plant on the first layer, producing a clear and consistent motion path.
    \item \textit{\cref{fig:multi_path}(d) Diamond Detection.} The model initially marks two candidate shapes that might satisfy the query. Through iterative refinement, the incorrect candidate fades; only the correct diamond remains circled in the end.
\end{itemize}

\subsubsection{Superposition-based Exploration.}
\label{sec:superposition}
Another distinctive mode observable along the diffusion trajectory is superposition-based exploration, where the model temporarily represents multiple mutually exclusive logical states simultaneously. Instead of committing early to a single configuration, the model maintains overlapping hypotheses that gradually resolve as noise is removed.
This phenomenon is particularly evident in tasks involving object reordering and spatial alignment\MO{, where alternative outcomes often occupy similar spatial regions and overlapping pixel regions, making superposed hypotheses directly visible.}




\begin{itemize}
    \item \textit{\cref{fig:multi_path}(e) Size Pattern Completion.} The size-pattern follows a repeating "large-medium–small" pattern. When predicting the next element, the model initially generates overlapping circles of different sizes, representing competing hypotheses about the correct continuation of the sequence.
    \item \textit{\cref{fig:multi_path}(f) Object Rotation.} Instead of a discrete rotation from one angle to another, the model produces a blurred superposition of multiple orientations.
\end{itemize}
\begin{figure}[t]
  \centering
  \includegraphics[width=\linewidth]{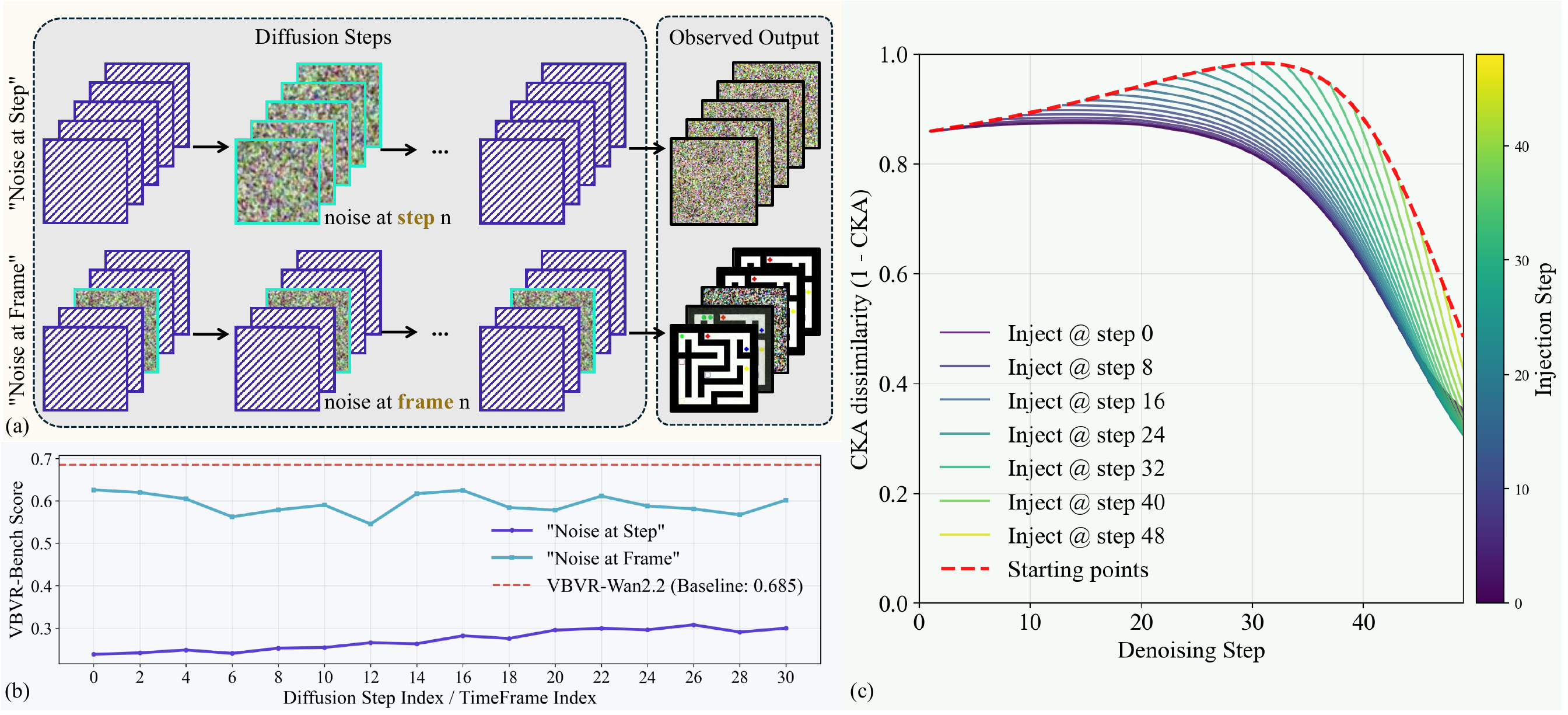}
  \caption{
    \textbf{Noise perturbation and information flow.} (a) Illustration of noise injection schemes; "Noise at Step" suffers more significant corruption than "Noise at Frame". (b) Performance drop with the two noise injection schemes. X-axis is the injection index (either diffusion step or frame). (c) Information flow across denoising steps (CKA dissimilarity: 1.0 indicates complete corruption, 0.0 indicates no effect).
  }
  \label{fig:noise_injection}
\end{figure}
\subsubsection{Discussion of Reasoning Patterns.}
\MO{We examine whether the observed \textit{multi-path exploration} and \textit{superposition-based exploration} reflect underlying reasoning patterns in video models.
\textbf{Prevalence.} To assess the prevalence of these phenomena, we manually inspect 200 examples drawn from VBVR-Bench and real-world videos. We find that 72\% exhibit either \textit{multi-path exploration} or \textit{superposition-based exploration} during denoising. The remaining examples typically involve simpler tasks whose trajectories converge directly to a solution.
\textbf{Validity.} We further argue that these phenomena cannot be fully attributed to generic diffusion artifacts or uncertainty. Conventional artifacts, such as blur and ghosting, generally lack clear semantic structure. In contrast, the early-stage candidates we observe are semantically coherent and task-consistent. 
Moreover, as shown in \cref{fig:multi_path}, we quantify the degree of intermediate exploration using \textit{candidate multiplicity}. Specifically, at each denoising step, we count the number of semantically distinct and task-consistent candidate solutions visible in the intermediate output. We find that this behavior is substantially more pronounced in the reasoning-enhanced VBVR-Wan2.2 model than in general video-generation baselines. 
Together, these findings suggest that the observed patterns reflect structured exploration beyond ordinary diffusion ambiguity.
}

\begin{figure}[!t]
    \centering
    \includegraphics[width=\textwidth]{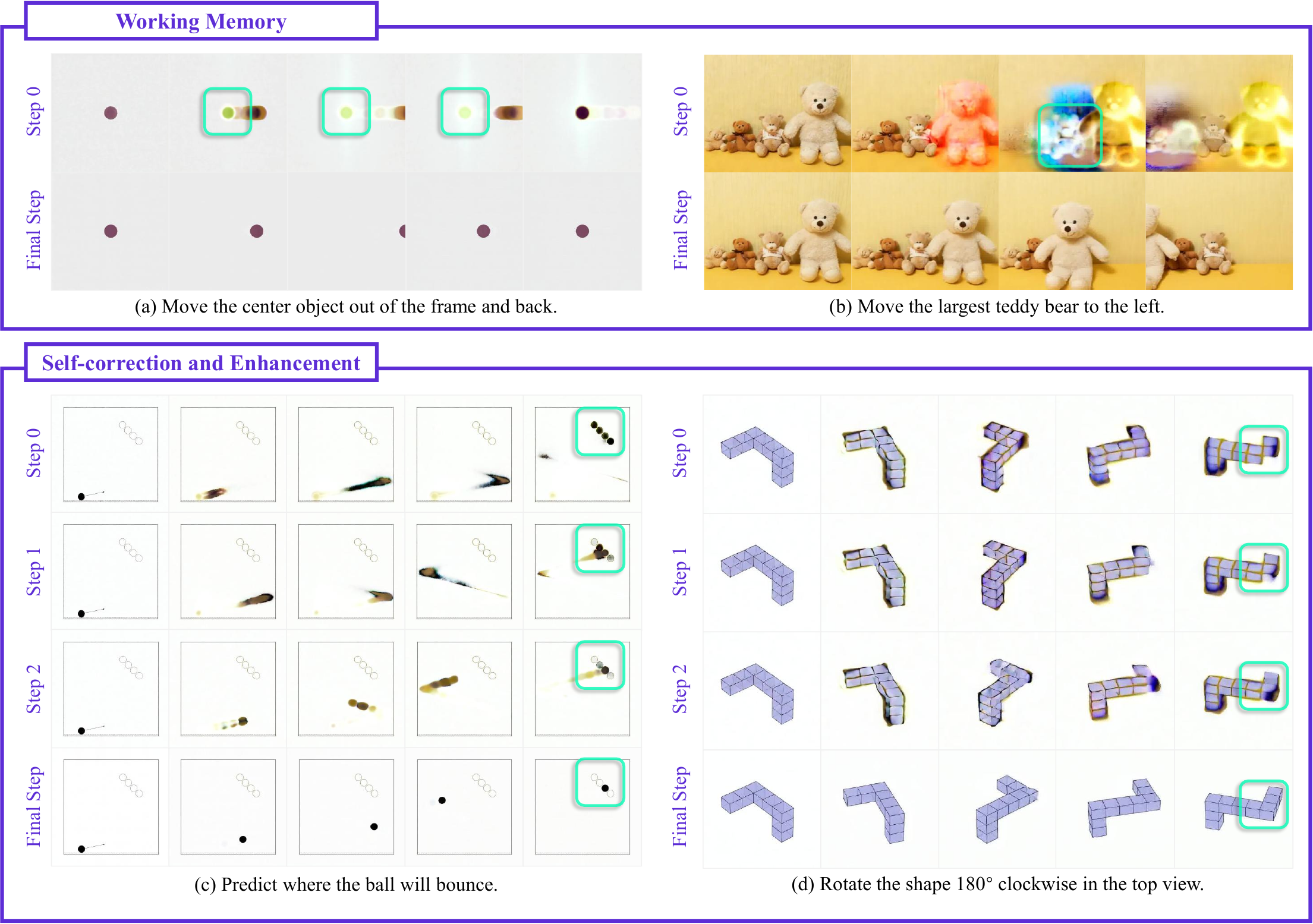}
    \caption{
    \textbf{Emergent reasoning behaviors: memory and self-correction.}
    (a) The center point is retained to guide the return motion.
    (b) The contour of the occluded small teddy bear is preserved, enabling the model to address object permanence.
    (c) The trajectory of the ball gradually extends and becomes complete.
    (d) The missing cube only appears in the later diffusion steps.
    Cyan boxes are added for illustration; they are not part of the generated video.
    }
    \label{fig:emergent_behaviors}
\end{figure}

\subsection{Noise Perturbation and Information Flow}
\label{sec:noise_perturbation}

Our hypothesis is further validated through targeted noise injection experiments. We compare two settings to isolate where the core reasoning process occurs:
1) "Noise at Step": $x_{s, \forall f} \leftarrow \mathcal{N}(0, \mathbf{I})$. That is, disruptive Gaussian noise is injected into all frames at a specific diffusion step.
2) "Noise at Frame": $x_{\forall s, f} \leftarrow \mathcal{N}(0, \mathbf{I})$. That is, Gaussian noise is injected into a specific frame across all diffusion steps.
The two settings are illustrated in~\cref{fig:noise_injection}(a).

In~\cref{fig:noise_injection}(b), we evaluate model performance under these two noise injection schemes. Compared to the baseline without noise, the "Noise at Step" setting causes the final score to collapse from 0.685 to below 0.3, indicating that the reasoning trajectory is highly sensitive to disruptions along the diffusion steps. Noise injected at a particular diffusion step therefore leads to a significant interruption of the model’s reasoning process.

In contrast, under "Noise at Frame" injection, the model demonstrates more robustness with a much smaller performance drop. This behavior can be explained by the architecture of diffusion transformers: each denoising step has full observation of the preceding latent sequence through bidirectional attention, allowing the model to refine the entire video latent jointly. Consequently, corrupted frames can be recovered by leveraging the uncorrupted information from neighboring frames during subsequent denoising steps.

In~\cref{fig:noise_injection}(c), we further analyze information propagation by measuring divergence after injecting noise at step $s_t$. We visualize CKA dissimilarity \cite{kornblith2019similarity}, where 1.0 indicates complete corruption and 0.0 indicates no effect. The results show that perturbations introduced in early diffusion steps propagate throughout the entire trajectory, fundamentally altering the final reasoning outcome. 

Moreover, the red dotted line highlights step-wise sensitivity to disruptive noise, which gradually increases and peaks around steps 20–30. This observation aligns with our qualitative analysis. Although steps 20–30 are not the earliest stages where we first observe reasoning phenomena, by this point the model has already pruned its reasoning trajectory toward the final conclusion. Consequently, perturbations at these steps have a large impact, as they can disrupt a reasoning process that is nearly finalized. Later steps, in contrast, appear less critical for the model’s reasoning capability.

%


\section{Emergent Reasoning Behaviors}

Similar to the emergent reasoning behaviors observed in Large Language Models (LLMs)~\cite{wei2022chain, madaan2023selfrefine, yao2023react, huang2023large, yang2025understanding}, we identify three surprising properties that are critical to effective video reasoning: \textit{working memory} (\cref{sec:memory}), which retains essential information throughout the reasoning process; \textit{self-correction and enhancement} (\cref{sec:self-correction}), which enables the model to revise intermediate hypotheses or refine previously generated answers, gradually adjusting toward the optimal solution even when it is not present initially; and \textit{perception before action} (\cref{sec:understanding_before_reasoning}), indicates that the model first establishes a grounded understanding of scene entities and spatial layout before reasoning about actions and interactions.
\subsection{Working Memory}
\label{sec:memory}

Reasoning requires the maintenance of "working memory" or a state. The demonstrations show that the diffusion process naturally establishes persistent anchors that preserve critical information across generation steps.

\begin{itemize}
    \item \textit{\cref{fig:emergent_behaviors}(a) Object Reappearance.} The model consistently preserves the object’s initial position throughout the diffusion steps, enabling the circle to return to its original location and remain consistent with the initial condition.
    \item \textit{\cref{fig:emergent_behaviors}(b) Teddy Bear Relocation.} During the movement task, the largest teddy bear temporarily blocks the small teddy bear on the left. Despite this occlusion, the early diffusion steps retain the state of the small bear to ensure consistent generation in the whole video.
\end{itemize}
\subsection{Self-correction and Enhancement}
\label{sec:self-correction}

During the diffusion process, we observe several stochastic “aha moments,” where the model initially selects an incorrect option but later revises its reasoning after a few diffusion steps, exploring an alternative strategy.
These behaviors are functionally analogous to the internal backtracking and "slow thinking" discussed in long-thinking Large Language Models (LLMs) \cite{yang2025understanding}. Importantly, such transitions are not limited to correcting mistakes. The model may also refine an initially incomplete answer into a logically richer and more comprehensive one, reflecting a form of latent self-improvement rather than simple error repair.

In contrast to the "Chain-of-Frames" theory, which would require such corrections to happen sequentially across time, these reversals take place globally across all frames simultaneously within a single diffusion step. This further provides strong evidence that the video generation model prioritizes global logical integrity over local, sequential frame-wise updates.

\begin{itemize}
    \item \textit{\cref{fig:emergent_behaviors}(c) Hit Target After Bounce.} Initially, the ball's trajectory is incomplete and ambiguous. As diffusion progresses, the model gradually completes the trajectory, making it increasingly clear, and the outcome converges from four candidate points to a single correct point.
    \item \textit{\cref{fig:emergent_behaviors}(d) 3D Shape Rotation.} At the first diffusion step, the rotated cubes are generated with incorrect quantities and arrangements. After several diffusion steps, the model gradually corrects both the number and the spatial configuration, producing a coherent and accurate final result.
\end{itemize}


\begin{figure}[!t]
\vspace{-4mm}
  \centering
  \includegraphics[width=\linewidth]{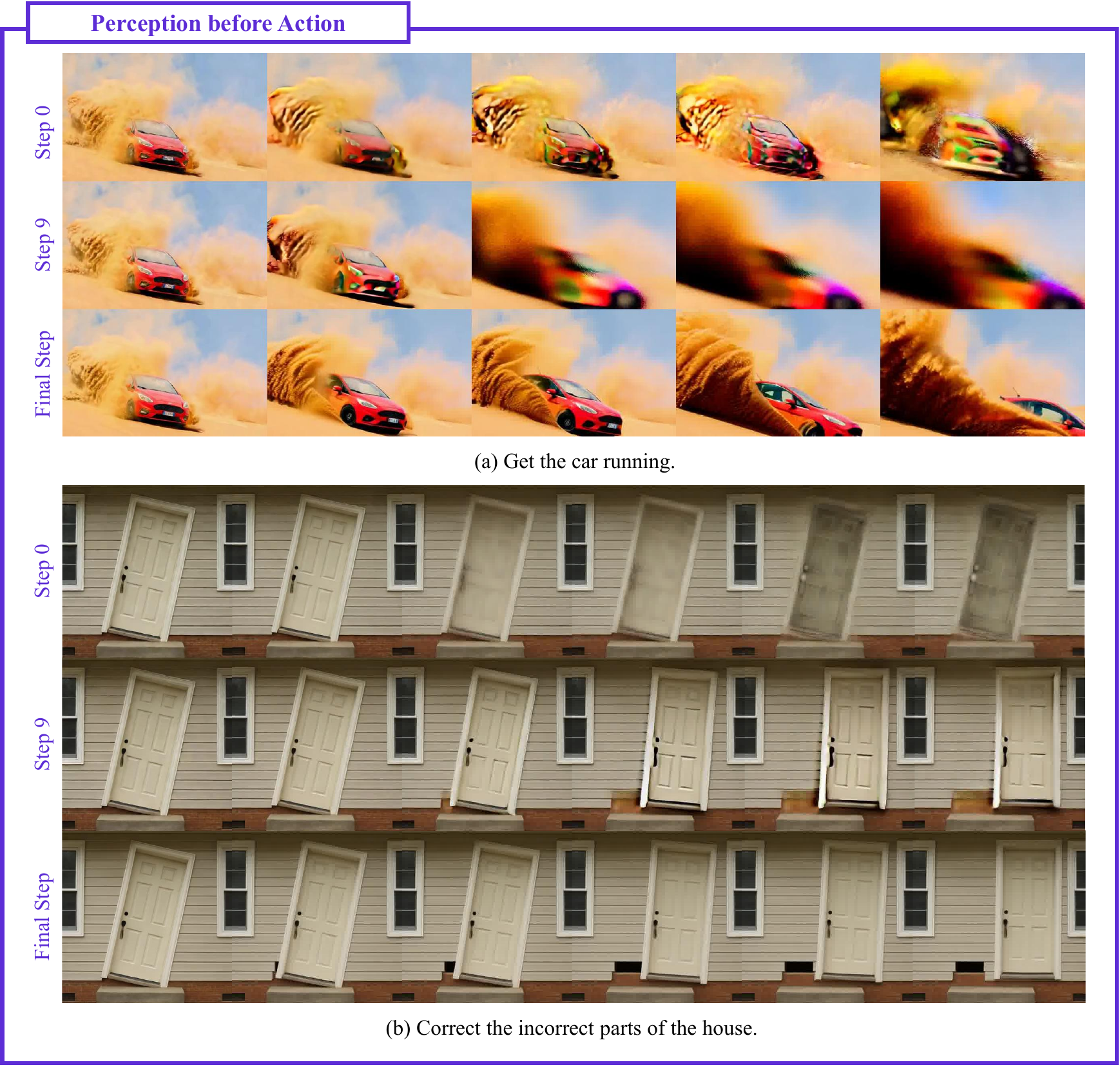}
  \caption{
  \textbf{Emergent reasoning behavior: understanding before reasoning.}
    (a) Early diffusion steps identify the car as the object of interest with no motion, while later steps introduce action and simulate physical interactions.
    (b) Early steps recognize the door as the target object with no motion, and later steps manipulate it.
  }
  \label{fig:understanding_viz}
\end{figure}

\subsection{Perception before Action}
\label{sec:understanding_before_reasoning}


We observe a phenomenon suggesting that the diffusion trajectory first addresses the "what" and "where" of a scene before determining the "how" and "why" of its thinking progression. This reflects a "Perception before Action" transition, where the model progresses from static grounding to dynamic reasoning. As illustrated in Fig.~\ref{fig:understanding_viz}, the initial diffusion step primarily identifies the foreground entity specified in the prompt  (\eg, the car or the door), without exhibiting explicit motion planning or relational transformation. Dynamic structure emerges only in later steps, as the model moves beyond scene grounding and starts coordinating object motion and inter-object interactions.

\section{Layer-wise Mechanistic Analysis}
\label{sec:layer-wise}
Inspired by the discovery of vision function layers in vision-language models~\cite{shi2025vision}, we investigate how diffusion transformers process visual information during video reasoning by analyzing the internal representations across transformer layers. Rather than focusing solely on generated outputs, we examine how hidden states evolve within the DiT backbone and how different layers contribute to semantic grounding and reasoning behaviors. Specifically, we study the model from two complementary perspectives. First, we visualize token-level activations across layers to analyze how representation energy distribute over spatial-temporal regions. Second, we perform layer-wise latent swapping experiments to causally evaluate how intermediate representations influence the final reasoning outcome. Together, these analyses provide a fine-grained view of how information is organized and progressively transformed inside the model.
\begin{figure}[t!]
  \centering
  \includegraphics[width=\linewidth]{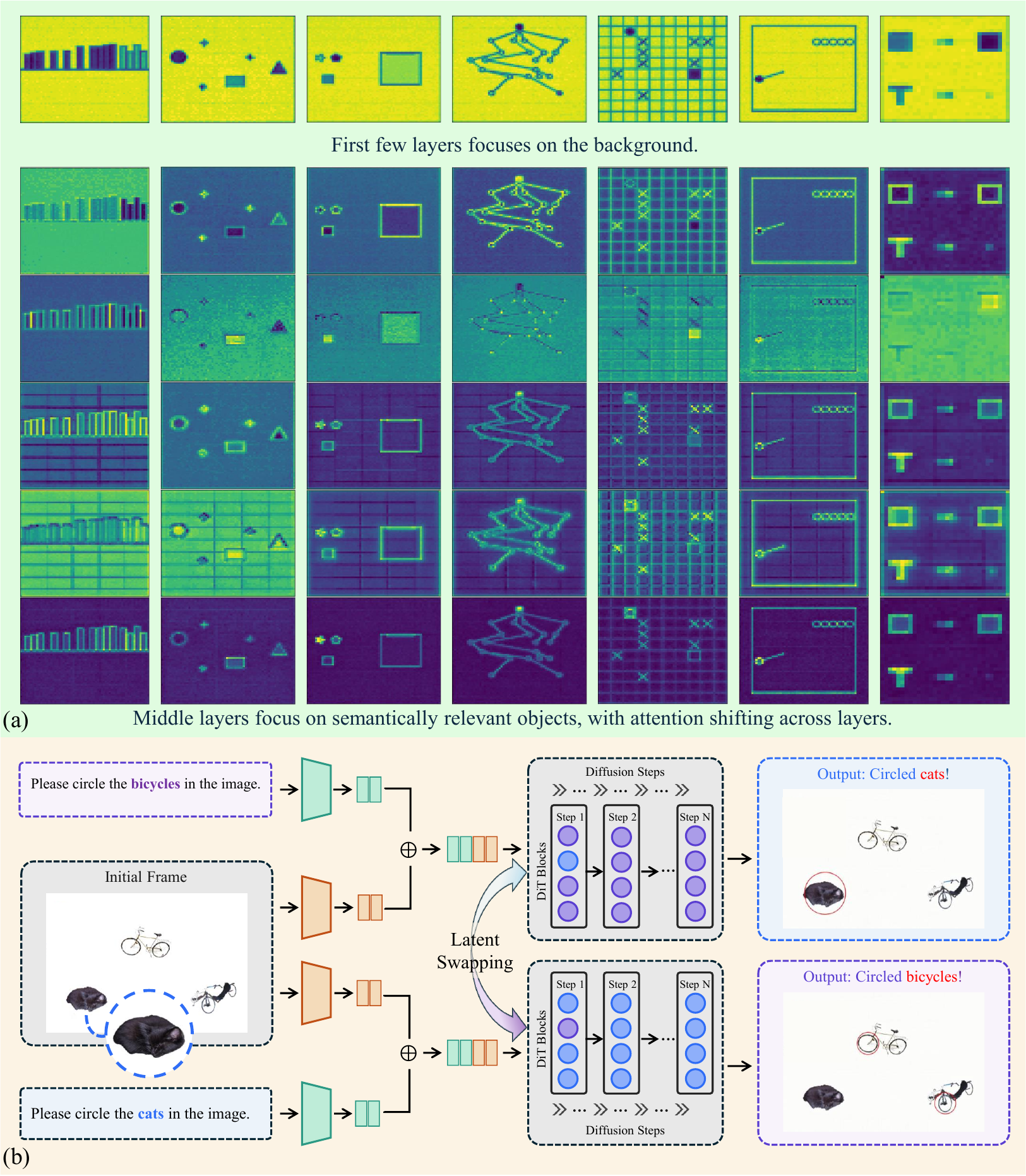}
  \setlength{\abovecaptionskip}{-6pt}
  \setlength{\belowcaptionskip}{-12pt}
  \caption{
    \textbf{Layer specialization.}
    (a) Layer-wise activation visualization shows that early layers of the video reasoning DiT tend to focus on background structures, whereas later layers carry out reasoning-related computations.
    (b) Layer-wise latent swapping reveals that certain middle layers (\eg, Layer 21 in this case) contain critical reasoning information that strongly influences the final outcome.  
  }
  \label{fig:understanding_before_reasoning}
\end{figure}
\subsection{Layer-wise Token-Level Visualization}
\label{sec:Layer-wise_vis}
To further investigate this transition, we analyze the internal activations of DiT blocks. During each diffusion step, forward hooks are registered on all transformer blocks of Wan2.2-I2V-A14B, and hidden states from the first forward pass are extracted to isolate the model's primary reasoning trajectory. The captured features are represented as a token sequence $feat \in \mathbb{R}^{B \times N \times D}$, where $N$ represents the total number of tokens and $D=5120$ is the embedding dimension. To restore the visual context, we utilize the grid dimensions $(f, h, w)$ captured from the model's \texttt{patch\_embedding} layer to reshape the features into a 5D tensor of shape $(B, f, h, w, D)$. Activation strength is then quantified by the channel-wise $L_2$ norm is computed along the channel dimension for each token, producing token-level energy maps that are visualized across layers and video frames. In \cref{fig:understanding_before_reasoning}(a), rows correspond to selected DiT layers (e.g., $L0, L10, L20 \dots L39$) and columns represent sequential video frames. Each cell in the grid displays a heatmap of token energies, enabling visualization of how the model's attention shifts throughout the network.

Within a single diffusion step, the earliest layers (Layers 0–9) primarily attend to global structures and background context. As computation proceeds through the layers of the same step, attention progressively shifts toward foreground entities and those specified in the prompt. From around Layer 9 onward, activations become increasingly concentrated on semantically relevant objects, accompanied by higher channel variance in localized regions corresponding to target entities. Notably, reasoning-related features also begin to emerge at this stage, with activations correlating with object motion and interactions. \MO{For example, in Column 4 of \cref{fig:understanding_before_reasoning}(a), the activation pattern evolves from the background to all nodes and paths, then narrows to the nodes themselves before revisiting the complete graph, Finally, the activation becomes concentrated on the starting node. This within-step progression is consistently observed across diffusion steps, indicating a recurrent hierarchy from global context to object-centric reasoning.}


\subsection{Layer-wise Latent Swapping Experiment}
As illustrated in \cref{fig:understanding_before_reasoning}(b), \MO{we perform a layer-wise latent swapping experiment to understand which transformer layers are responsible for determining the final grounding result. The experiment is performed on an object grounding task at the first diffusion step using a controlled setup of blank background and two object configurations ($O_A, O_B$). Two forward passes are first obtained using $O_A$ and $O_B$, producing latent representations $U^{(l)}_A$ and $U^{(l)}_B$ at each transformer layer. Then, hybrid forward passes for each layer k are constructed by replacing only the representation at layer k with that from the alternative configuration while keeping all other layers unchanged}: $\tilde{U}^{(k)} \leftarrow U_{alt}^{(k)}$, \text{and} $U^{(l \neq k)} = U_{orig}^{(l)}$.

\MO{Experiments are conducted on ten additional object categories across five seeds. Remarkably, swapping representations at one middle layer (around 15 to 35)} is sufficient to reverse the grounding result, causing the predicted target object to flip. This suggests that middle-to-late layers contain semantically decisive information for object grounding. \MO{A broader layer-wise analysis of flip rates is provided in Appendix \cref{sec:flip_rate}.}
 
\section{Training-Free Ensemble}
\begin{table*}[t]
\centering
\scriptsize
\setlength{\tabcolsep}{3pt}
\caption{Benchmarking results on VBVR-Bench. Overall In-Domain (ID) and Out-of-Domain (OOD) scores are reported alongside category-wise performance. Higher is better. \textbf{Bold}: best in group; \underline{underline}: second best.}
\resizebox{1.0\linewidth}{!}{
\begin{tabular}{l|c|c|ccccc|c|ccccc}
\toprule
& \multicolumn{1}{c|}{\textbf{}} 
& \multicolumn{6}{c|}{\textbf{In-Domain by Category}} 
& \multicolumn{6}{c}{\textbf{Out-of-Domain by Category}} \\
\cmidrule(lr){3-3}
\cmidrule(lr){4-8}
\cmidrule(lr){9-9}
\cmidrule(lr){10-14}
\textbf{Models}
& \textbf{Overall}
& \textbf{Avg.}
& \textbf{Abst.} & \textbf{Know.} & \textbf{Perc.} & \textbf{Spat.} & \textbf{Trans.}
& \textbf{Avg.}
& \textbf{Abst.} & \textbf{Know.} & \textbf{Perc.} & \textbf{Spat.} & \textbf{Trans.} \\
\midrule
\textbf{Human} 
& 0.974 & 0.960 & 0.919 & 0.956 & 1.00 & 0.95 & 1.00 & 0.988
& 1.00 & 1.00 & 0.990 & 1.00 & 0.970 \\
\midrule
\rowcolor{line-blue}\textbf{Open-source Video Models} & & & & & & & & & & & & & \\

CogVideoX1.5-5B-I2V~\cite{yang2024cogvid} 
& 0.273 & 0.283 & 0.241 & 0.328 & 0.257 & 0.328 & 0.305 
& 0.262 & \underline{0.281} & 0.235 & 0.250 & \textbf{0.254} & 0.282 \\ 

HunyuanVideo-I2V~\cite{kong2024hunyuan} 
& 0.273 & 0.280 & 0.207 & 0.357 & 0.293 & 0.280 & \underline{0.316} 
& 0.265 & 0.175 & \textbf{0.369} & 0.290 & \underline{0.253} & 0.250 \\ 

Wan2.2-I2V-A14B~\cite{wan_wan2025wan} 
& \textbf{0.371} & \textbf{0.412} & \textbf{0.430} & \textbf{0.382} & \textbf{0.415} & \textbf{0.404} & \textbf{0.419} 
& \textbf{0.329} & \textbf{0.405} & 0.308 & \textbf{0.343} & 0.236 & \underline{0.307} \\ 

LTX-2~\cite{hacohen2026ltx2} 
& \underline{0.313} & \underline{0.329} & \underline{0.316} & \underline{0.362} & \underline{0.326} & \underline{0.340} & 0.306 
& \underline{0.297} & 0.244 & \underline{0.337} & \underline{0.317} & 0.231 & \textbf{0.311} \\

\midrule
\rowcolor{line-blue}\textbf{Proprietary Video Models} & & & & & & & & & & & & & \\

Runway Gen-4 Turbo~\cite{runway_gen4_2025} 
& 0.403 & 0.392 & 0.396 & 0.409 & 0.429 & 0.341 & 0.363 
& 0.414 & 0.515 & \underline{0.429} & 0.419 & 0.327 & 0.373 \\ 

Sora 2~\cite{sora_openai_2025} 
& \textbf{0.546} & \textbf{0.569} & \underline{0.602} & \underline{0.477} & \textbf{0.581} & \textbf{0.572} & \textbf{0.597} 
& \textbf{0.523} & \underline{0.546} & \textbf{0.472} & \textbf{0.525} & \textbf{0.462} & \textbf{0.546} \\ 

Kling 2.6~\cite{kling_ai_kuai_2025} 
& 0.369 & 0.408 & 0.465 & 0.323 & 0.375 & 0.347 & \underline{0.519} 
& 0.330 & 0.528 & 0.135 & 0.272 & 0.356 & 0.359 \\ 

Veo 3.1~\cite{veo3.1_deepmind_2026} 
& \underline{0.480} & \underline{0.531} & \textbf{0.611} & \textbf{0.503} & \underline{0.520} & \underline{0.444} & 0.510 
& \underline{0.429} & \textbf{0.577} & 0.277 & \underline{0.420} & \underline{0.441} & \underline{0.404} \\ 

\midrule
\rowcolor{line-blue}\textbf{Video Reasoning Models} & & & & & & & & & & & & & \\

VBVR-Wan2.2~\cite{vbvr2026}
& \underline{0.685} & \underline{0.760} & \underline{0.724} & \textbf{0.750} & \underline{0.782} & \underline{0.745} & \underline{0.833} 
& \underline{0.610} & \underline{0.768} & \underline{0.572} & \textbf{0.547} & \underline{0.618} & \underline{0.615} \\ 

VBVR-Wan2.2 + Training-Free Ensemble & \textbf{0.716} & \textbf{0.780} & \textbf{0.760} & \underline{0.744} & \textbf{0.809} & \textbf{0.749} & \textbf{0.858} & \textbf{0.650} & \textbf{0.803} & \textbf{0.705} & \underline{0.531} & \textbf{0.639} & \textbf{0.716} \\
\bottomrule
\end{tabular}
}
\label{tab:vbvr_results}
\end{table*}

To further validate our key observations (\eg, multi-path reasoning during the early diffusion steps in~\cref{sec:cos} and~\cref{sec:understanding_before_reasoning}, and reasoning-active features emerge in the mid-layers in~\cref{sec:layer-wise}), we design a simple proof-of-concept method, which we call Training-free Ensemble (TFE).
Even simpler than merging models within the same optimization basin~\cite{2022modelsoups}, we implement a multi-seed ensemble of the same model at the latent level during \textit{early} diffusion steps, with the hypothesis that the reasoning manifold (the internal latent space) contains a shared probabilistic bias toward the correct outcome.
Specifically, we execute three independent forward passes using different initial noise seeds. During the first diffusion step ($s=0$), we extract the hidden representations $U^{(l)}$ from the transformer backbone, and perform a spatial-temporal averaging of the latents across mid-layers: 20 to 29. 
By aggregating representations within this specific reasoning window, we effectively perform a latent-space ensemble resembling expert voting. This operation filters out seed-specific noise and biases the model’s probability distribution toward a more stable and logically consistent latent state.

\begin{wraptable}[8]{l}{0.4\linewidth}
\setlength{\intextsep}{0pt}
\setlength{\floatsep}{0pt}
\setlength{\textfloatsep}{0pt}
\vspace{-4mm}
\footnotesize

\setlength{\abovecaptionskip}{-0pt}
\centering
\caption{TFE results on four benchmarks}
\label{tab:training-free-more-benchmark}
\setlength{\tabcolsep}{2pt}
\begin{tabular}{lcccc}
\hline
\textbf{Method} & 
\shortstack{\textbf{VBVR-}\\\textbf{Bench}} & 
\shortstack{\textbf{V-Reason}\\\textbf{Bench}} & 
\shortstack{\textbf{MME-}\\\textbf{CoF}} & 
\shortstack{\textbf{RISE-}\\\textbf{Video}} \\
\hline
VBVR-Wan2.2 & 0.69 & 8.94 & 1.30 & 61.60 \\
 + TFE & 0.72 & 12.12 & 1.52 & 65.85 \\
VBVR-Wan2.1 & 0.59 & 12.66 & 0.36 & 37.35 \\
 + TFE & 0.61 & 13.35 & 0.38 & 42.92 \\
VBVR-LTX2.3 & 0.52 & 3.61 & 0.12 & 32.75 \\
 + TFE & 0.53 & 3.19 & 0.27 & 35.85 \\
\hline
\end{tabular}
\vspace{-2mm}
\end{wraptable}
 
\MO{\cref{tab:vbvr_results} shows that TFE improves VBVR-Wan2.2 across sub-categories in VBVR-Bench~\cite{vbvr2026}. Extending the evaluation to multiple video generation base models (LTX2.3, Wan2.1, and Wan2.2) across four benchmarks (VBVR-Bench, V-Reason Bench~\cite{v-reason}, MME-CoF~\cite{mme-cof}, and RISE-Video~\cite{rise-video}) shows similar trends in \cref{tab:training-free-more-benchmark}, where TFE consistently improves performance across nearly all model-benchmark pairs without any additional training. For instance, VBVR-Wan2.2  achieves relative improvements of 4.3\% on VBVR-Bench, 35.6\% on V-Reason Bench, 16.9\% on MME-CoF, and 6.9\% on RISE-Video. Similar improvements are observed for VBVR-Wan2.1 and VBVR-LTX2.3, demonstrating that the proposed latent-space aggregation strategy is model-agnostic and transfers across different video diffusion backbones.}




\section{Conclusion}

In this work, we investigate the mechanisms underlying reasoning in diffusion-based video generation models. Contrary to the previously hypothesized Chain-of-Frames (CoF) mechanism, we show that reasoning primarily unfolds along diffusion steps, which we term Chain-of-Steps (CoS), through qualitative analysis and targeted perturbation experiments. Our study further reveals several emergent reasoning behaviors, including working memory retention, self-correction, and perception before action.  Analysis of Diffusion Transformer layers shows that the middle layers are important for reasoning. To further validate these insights, we propose a simple training-free reasoning path ensemble method that achieves performance improvements over a strong baseline. 

%
%

\newpage
\bibliographystyle{splncs04}
\bibliography{main}
\newpage
\beginappendix
\appendix

\section{More Experiments on Training-Free Ensemble}

\subsection{Experiment on ensemble size}
\begin{wraptable}[8]{l}{0.3\linewidth}
\setlength{\intextsep}{0pt}
\setlength{\floatsep}{0pt}
\setlength{\textfloatsep}{0pt}
\vspace{-2mm}
\footnotesize
\setlength{\abovecaptionskip}{-0pt}
\centering
\caption{TFE scaling curve}
\label{tab:seed_ablation}
\setlength{\tabcolsep}{3pt}
\setlength{\aboverulesep}{0pt}
\setlength{\belowrulesep}{0pt}

\begin{tabular}{lccc}
\toprule
\textbf{\# Seeds}
& \textbf{Overall}
& \textbf{ID}
& \textbf{OOD} \\
\midrule
1 & 0.685 & 0.760 & 0.610 \\
2 & 0.688 & 0.770 & 0.606 \\
3 (TFE default)
& \underline{0.716}
& \underline{0.780}
& \underline{0.650} \\
4 & \textbf{0.729} & 0.781 & \textbf{0.676} \\
5 & 0.720 & \textbf{0.782} & 0.658 \\
6 & 0.675 & 0.771 & 0.579 \\
\bottomrule
\end{tabular}
\end{wraptable}
\MO{We further investigate the effect of ensemble size by varying the number of seeds from 2 to 6 (\cref{tab:seed_ablation}). Performance improves as additional seeds introduce}\MO{complementary reasoning trajectories, reaching its best performance with 3--4 seeds. Increasing the ensemble size further leads to a performance drop, indicating that excessive averaging may introduce conflicting reasoning paths and weaken useful signals. This finding suggests a trade-off between diversity and consistency in latent-space aggregation. Three seeds are chosen for TFE as a computationally efficient default setting.}

\subsection{Experiment on Aggregation window}
\begin{table*}[!h]
\centering
\scriptsize
\setlength{\tabcolsep}{3pt}
\caption{Comparison of VBVR-Bench performance across different layer windows at diffusion step $s=0$. Mid-layer aggregation (20–29) achieves the best overall performance (0.716) by capturing the critical reasoning-active window. \textbf{Bold}: best in group; \underline{underline}: second best.}
\resizebox{1.0\linewidth}{!}{
\begin{tabular}{l|c|c|ccccc|c|ccccc}
\toprule
& \multicolumn{1}{c|}{\textbf{}} 
& \multicolumn{6}{c|}{\textbf{In-Domain by Category}} 
& \multicolumn{6}{c}{\textbf{Out-of-Domain by Category}} \\
\cmidrule(lr){3-3}
\cmidrule(lr){4-8}
\cmidrule(lr){9-9}
\cmidrule(lr){10-14}
\textbf{Aggregated Layers}
& \textbf{Overall}
& \textbf{Avg.}
& \textbf{Abst.} & \textbf{Know.} & \textbf{Perc.} & \textbf{Spat.} & \textbf{Trans.}
& \textbf{Avg.}
& \textbf{Abst.} & \textbf{Know.} & \textbf{Perc.} & \textbf{Spat.} & \textbf{Trans.} \\

\midrule

\rowcolor{line-blue}baseline(VBVR-Wan2.2)~\cite{vbvr2026}
& 0.685 & 0.760 & 0.724 & \textbf{0.750} & 0.782 & 0.745 & 0.833
& 0.610 & 0.768 & 0.572 & \textbf{0.547} & 0.618 & 0.615 \\ 
0-9  & 0.688 & \underline{0.774} & \underline{0.754} & 0.741 & \underline{0.808} & 0.746 & \underline{0.835} & 0.602 & 0.805 & \underline{0.635} & 0.485 & 0.597 & 0.642 \\ 
0-39 & \underline{0.690} & 0.767 & 0.733 & 0.737 & 0.807 & \underline{0.747} & 0.825 & \underline{0.613} & \textbf{0.830} & 0.606 & 0.482 & \underline{0.630} & \underline{0.657} \\ 
20-29(Training-Free Ensemble) & \textbf{0.716} & \textbf{0.780} & \textbf{0.760} & \underline{0.744} & \textbf{0.809} & \textbf{0.749} & \textbf{0.858} & \textbf{0.650} & \underline{0.803} & \textbf{0.705} & \underline{0.531} & \textbf{0.639} & \textbf{0.716} \\
\bottomrule
\end{tabular}
}
\label{tab:supp_ablation}
\end{table*}

To examine the impact of the aggregation window, we conduct an experiment over different layer ranges when performing the latent ensemble. For a fair comparison, all variants perform the ensemble at the first diffusion step ($s=0$), while only the aggregated layer window is varied. As shown in \cref{tab:supp_ablation}, aggregating representations from the early layers (0–9) yields only marginal improvement over the baseline, increasing the overall score from 0.685 to 0.688, with limited gains in both in-domain and out-of-domain settings. This suggests that early-layer representations primarily encode low-level perceptual features and have not yet formed the semantic structures required for reasoning. Expanding the aggregation to all layers (0–39) produces a slightly higher overall score (0.690), but the improvement remains modest and inconsistent across categories, indicating that averaging across the entire depth introduces noise from layers that are either too early (perceptual) or too late (already specialized for generation). In contrast, aggregating mid-layer representations (layers 20–29) achieves the best performance, reaching an overall score of 0.716 and consistently improving most categories. This result aligns with our earlier analysis that the middle layers correspond to the transition stage between understanding and reasoning, where the model actively integrates semantic concepts and forms reasoning trajectories. Consequently, performing the ensemble within this reasoning-active window provides a more stable and informative latent representation, effectively filtering stochastic noise across seeds while preserving the semantic structures that guide correct reasoning.

\section{Additional Layer-wise Mechanistic Analysis}
\subsection{Flip Rate Analysis of Layer-wise Swapping Experiments}
\begin{figure*}[t!]
    \vspace*{-2mm}
    \centering
    \includegraphics[width=\textwidth]{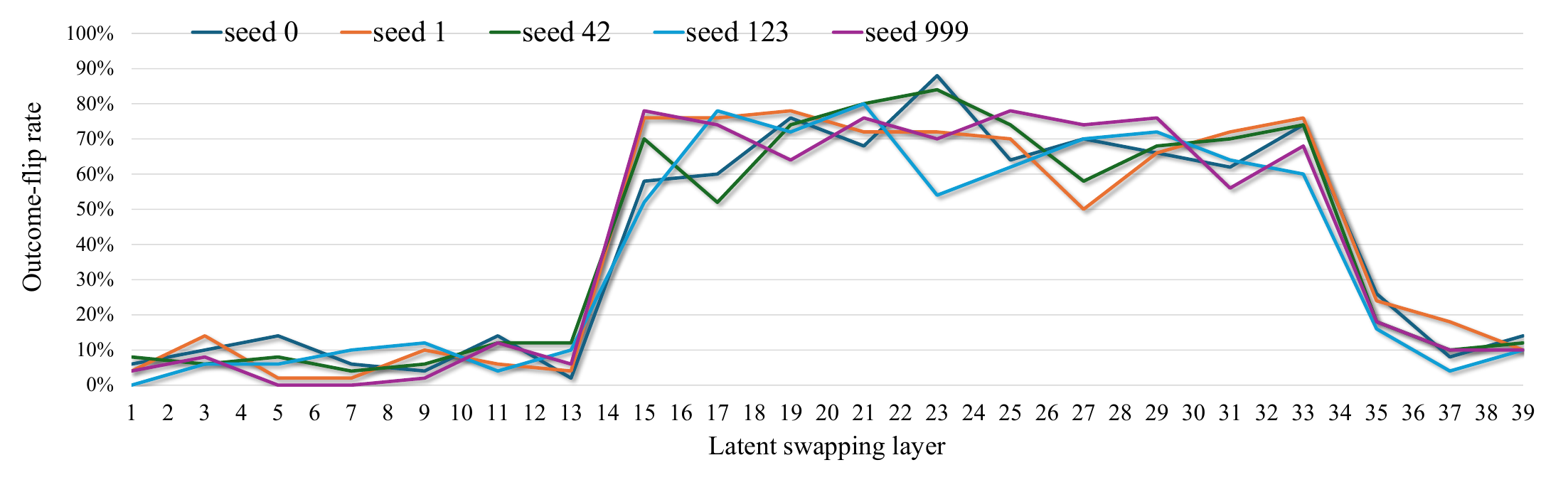}
    \setlength{\abovecaptionskip}{-6pt}
    \caption{The outcome-flip rates of layer-wise swapping experiments always peak at middle layers across different seeds}
    \setlength{\belowcaptionskip}{-6pt}
    \vspace*{-6mm}
    \label{fig:flip_rate}
\end{figure*}
\label{sec:flip_rate}
\MO{The layer-wise swapping experiments are conducted on ten object-category pairs (cat vs. bicycle, cat vs. dog, cat vs. horse, cat vs. cow, dog vs. bicycle, dog vs. cow, dog vs. sheep, circle vs. square, circle vs. triangle, and triangle vs. square), each evaluated across five random seeds. As shown in \cref{fig:flip_rate}, we consistently observe a characteristic bell-shaped pattern in the resulting flip-rate curves. The flip rate remains relatively low in early layers, rises sharply in the middle layers, and then decreases again toward the final layers. This trend holds across all category pairs and random seeds, indicating that the phenomenon is robust rather than task-specific. The elevated flip rates in the middle layers suggest that these layers are primarily responsible for semantic decision-making, where competing hypotheses remain susceptible to perturbations. These findings provide additional evidence that semantic-related grounding are concentrated in the middle portion of the network.}
\subsection{Full Layer-wise Analysis}
This section presents a comprehensive visualization of token activation energies across all 40 DiT blocks and video frames, covering all tasks reported in \cref{sec:Layer-wise_vis} and \cref{fig:understanding_before_reasoning}. Each row corresponds to a transformer block, and each column corresponds to a video frame. The heatmaps show the spatial distribution of token activation magnitudes. In \cref{sec:Layer-wise_vis}, we discuss a layer-wise transition in which early layers focus on global structures, while middle layers increasingly attend to prompt-relevant foreground objects and exhibit reasoning-related features associated with object motion and interactions. We discuss two additional findings. 

\subsubsection{High sparsity in token activations.} Across most layers, a large fraction of spatial tokens exhibit very low activation norms (dark purple regions). This indicates that only a small subset of tokens carry significant signal at any given layer. In practice, this suggests that the model performs highly sparse computation in token space, where meaningful reasoning is concentrated in localized patches corresponding to salient visual structures. Such sparsity becomes particularly pronounced in middle layers, where entire spatial regions remain near-zero while only a few tokens remain strongly active.

\subsubsection{High concentration of token activations in the middle layers.} Beginning in the intermediate layers, regular grid-like patterns become clearly visible. These patterns might align with the underlying patch tokenization structure of the transformer, potentially indicating spatial awareness. At this stage, the model may begin organizing information along patch boundaries, resulting in checkerboard- or lattice-like activation patterns.

\begin{figure}[p]
\label{fig:feature_viz}
\centering
\includegraphics[width=\textwidth]{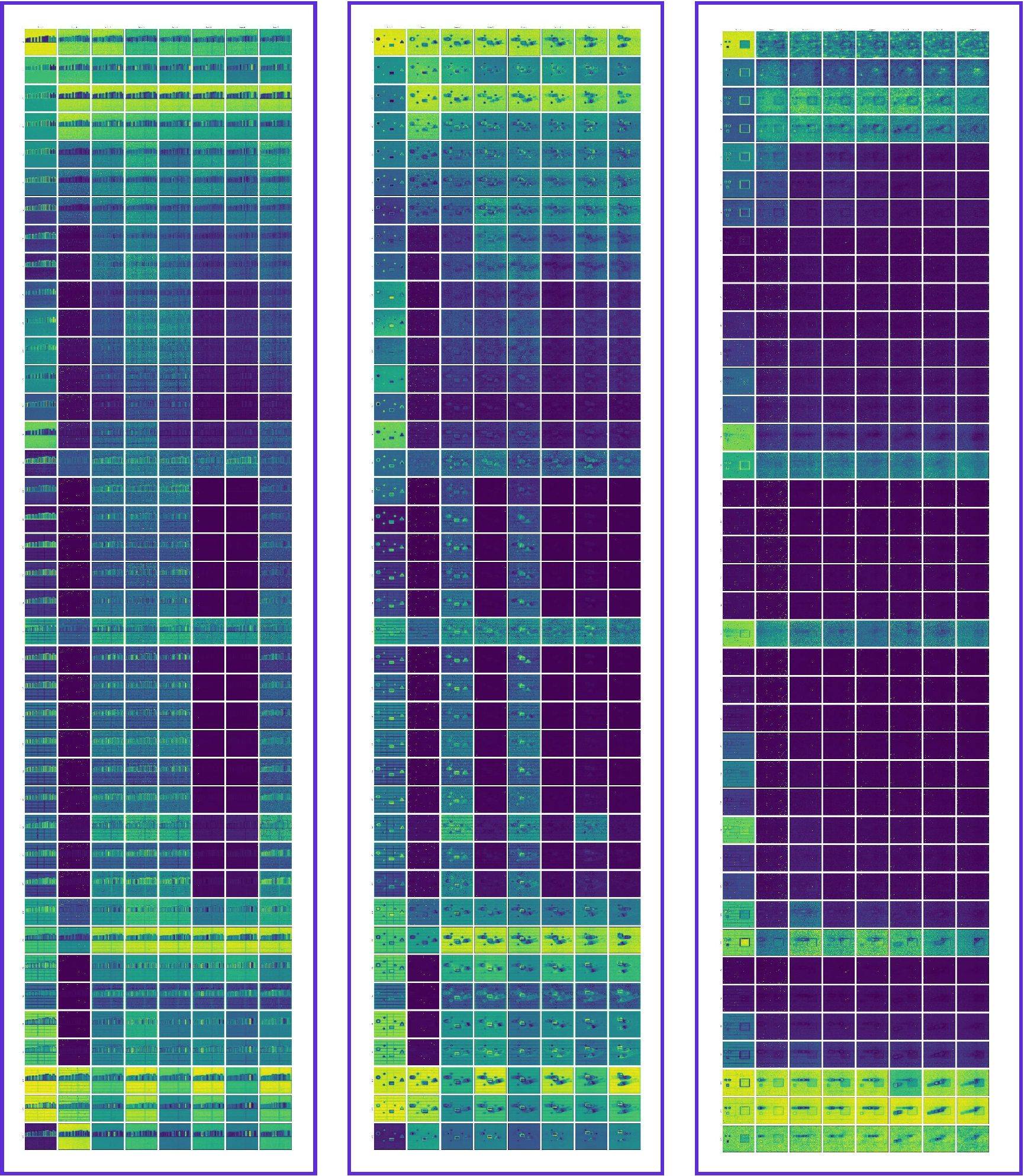}
\caption{Layer-wise token activation visualization across all 40 DiT blocks. Rows correspond to layers 0–39 (from top to bottom), while columns represent video frames.}
\end{figure}
\begin{figure}[p]\ContinuedFloat
\centering
\includegraphics[width=\textwidth]{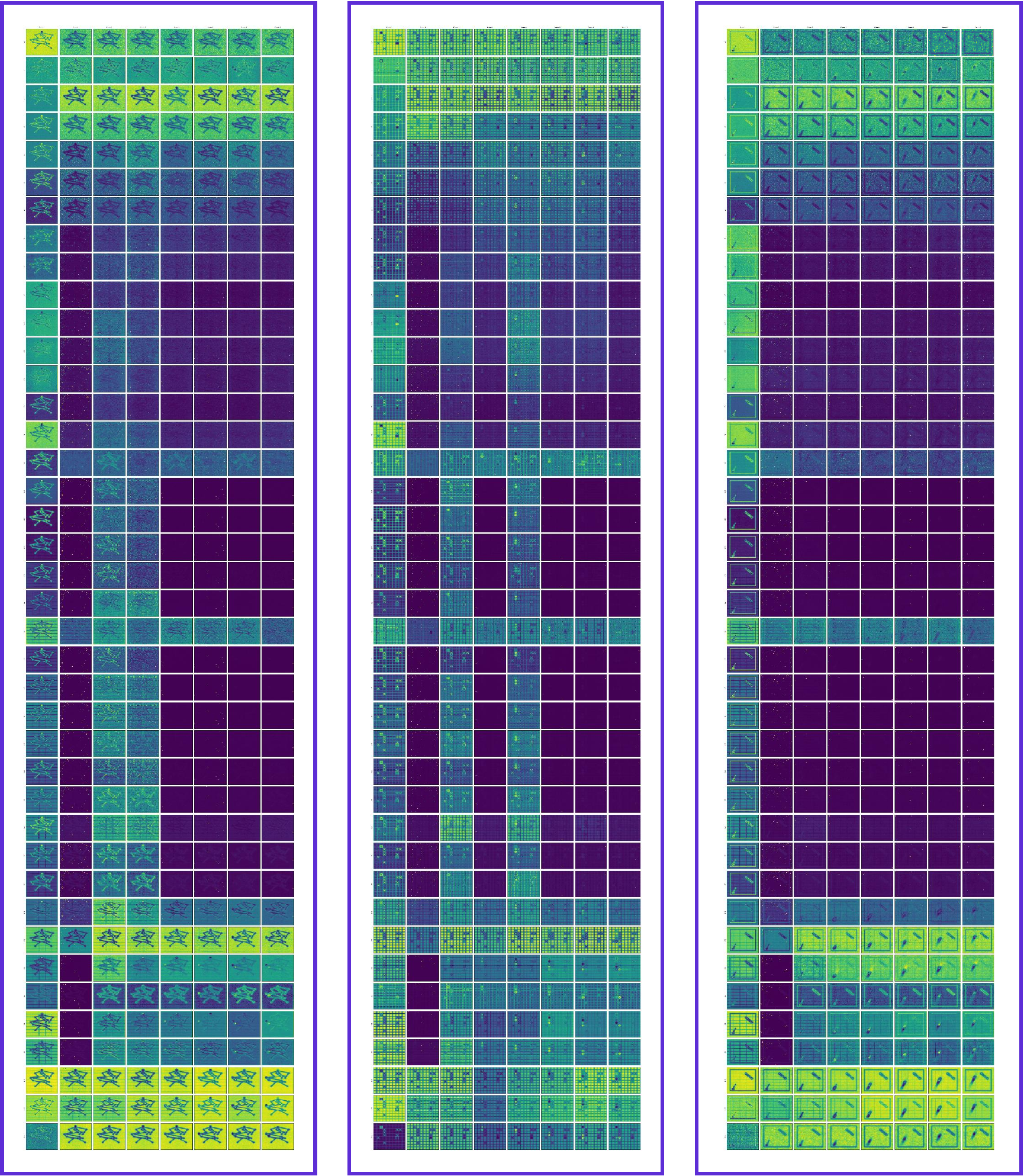}
\caption{(Continued) Layer-wise token activation visualization across all 40 DiT blocks. Rows correspond to layers 0–39 (from top to bottom), while columns represent video frames.}
\end{figure}

\section{The Impact of the Number of Frames}
\begin{table}[!h]
\centering
\caption{Comparison of Model Performance Across Frame Counts. Chronoedit here could be considered as an single-frame version of VBVR-Wan2.2. The original VBVR-Wan2.2 operates on $\sim$100 frames on average. }
\label{tab:reduce_frame}
\begin{tabular}{lccc}
\toprule
\textbf{Configuration} & \textbf{Overall} & \textbf{In-Domain} & \textbf{Out-of-Domain} \\ 
\midrule
Chronoedit & 0.581 & 0.637 & 0.524 \\ 
\midrule
5 frames  & 0.619 & 0.688 & 0.549 \\
9 frames  & 0.632 & 0.716 & 0.549 \\
17 frames & 0.663 & 0.743 & 0.582 \\
33 frames & 0.685 & 0.685 & 0.685 \\
65 frames & 0.675 & 0.760 & 0.591 \\ 
\midrule
VBVR-Wan2.2 & \textbf{0.685} & \textbf{0.760} & \textbf{0.610} \\ \bottomrule
\end{tabular}
\end{table}

Although \cref{sec:cos} shows that reasoning primarily occurs across diffusion steps rather than across frames, we observe that the number of frames still plays an important role. In practice, frames serve as a latent spatiotemporal workspace (or ``scratchpad'') that enables the diffusion model to store essential visual information throughout the diffusion process.

In \cref{tab:reduce_frame}, following ChronoEdit~\cite{wu2025chronoedit}, we conduct an experiment that repurposes the video generation model as an image editing model to simulate \textit{single-frame} reasoning. Specifically,, the 3D-factorized Rotary Position Embedding (RoPE)~\cite{su2021roformer} is modified to anchor the input image at time step 0 and the output image at a predefined time step $T$, while intermediate frames are dropped after a few steps. However, this configuration performs substantially worse than all \textit{multi-frame} settings, suggesting that maintaining multiple frames is essential for capturing spatiotemporal coherence and enabling effective video reasoning.

We further investigate the effect of reducing the number of generated frames in VBVR-Wan2.2. The performance drop is relatively minor when the number of frames decreases from the original $ \sim 100$ to around 17. However, further reducing the frame count leads to noticeable degradation. This observation reinforces our hypothesis that although reasoning does not occur strictly in a frame-wise manner, maintaining a minimum level of temporal continuity is still necessary to accommodate key events required for correct inference.

\section{Performance on 4-Step Distilled Model}

We investigate how distillation affects reasoning in video generation models using a distilled 4-step Wan2.2-I2V-14B model. Distillation significantly compresses the denoising trajectory, raising the question of whether the reasoning dynamics remain observable under such a shortened inference process. To study this, we simultaneously adapt two LoRA models with scaling weights of 0.5 each: one based on VBVR-Wan2.2 that enhances reasoning capability, and the other based on a 4-step model distilled via Phased DMD~\cite{fan2025phased} that improves generation speed.

We find that although the number of denoising steps is drastically reduced from 50 to 4, the steps required for reasoning cannot be compressed proportionally. In particular, the characteristic reasoning activity that typically emerges during the early diffusion steps persists even after distillation. However, we also observe that in some tasks the noise scheduler reduces the noise level too aggressively in the first step, collapsing the latent exploration phase where reasoning signals usually emerge. As a result, intermediate reasoning patterns become difficult to observe, and the overall performance on VBVR-Bench drops significantly from 0.685 to 0.605.
\begin{figure}[!b]
\centering
\includegraphics[width=0.95\textwidth]{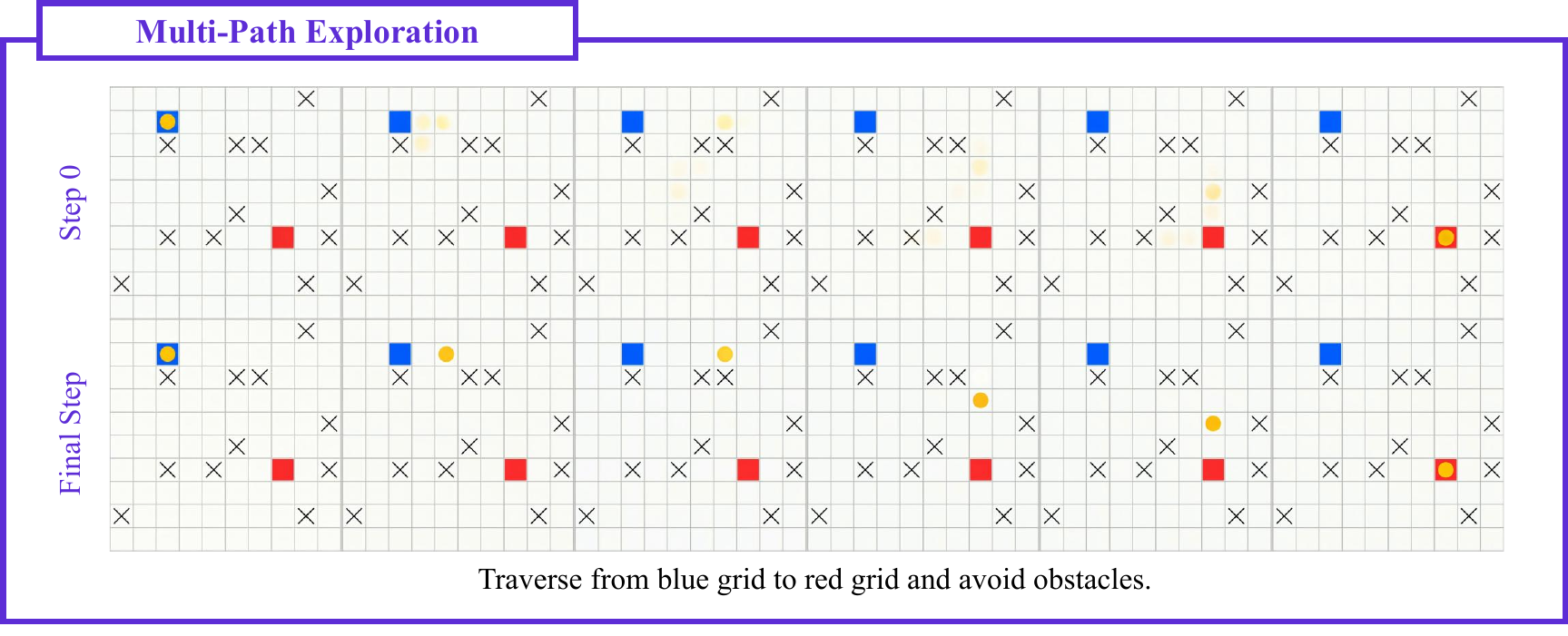}
\end{figure}
\begin{figure}[!t]
\centering
\vspace{-8pt}
\includegraphics[width=0.95\textwidth]{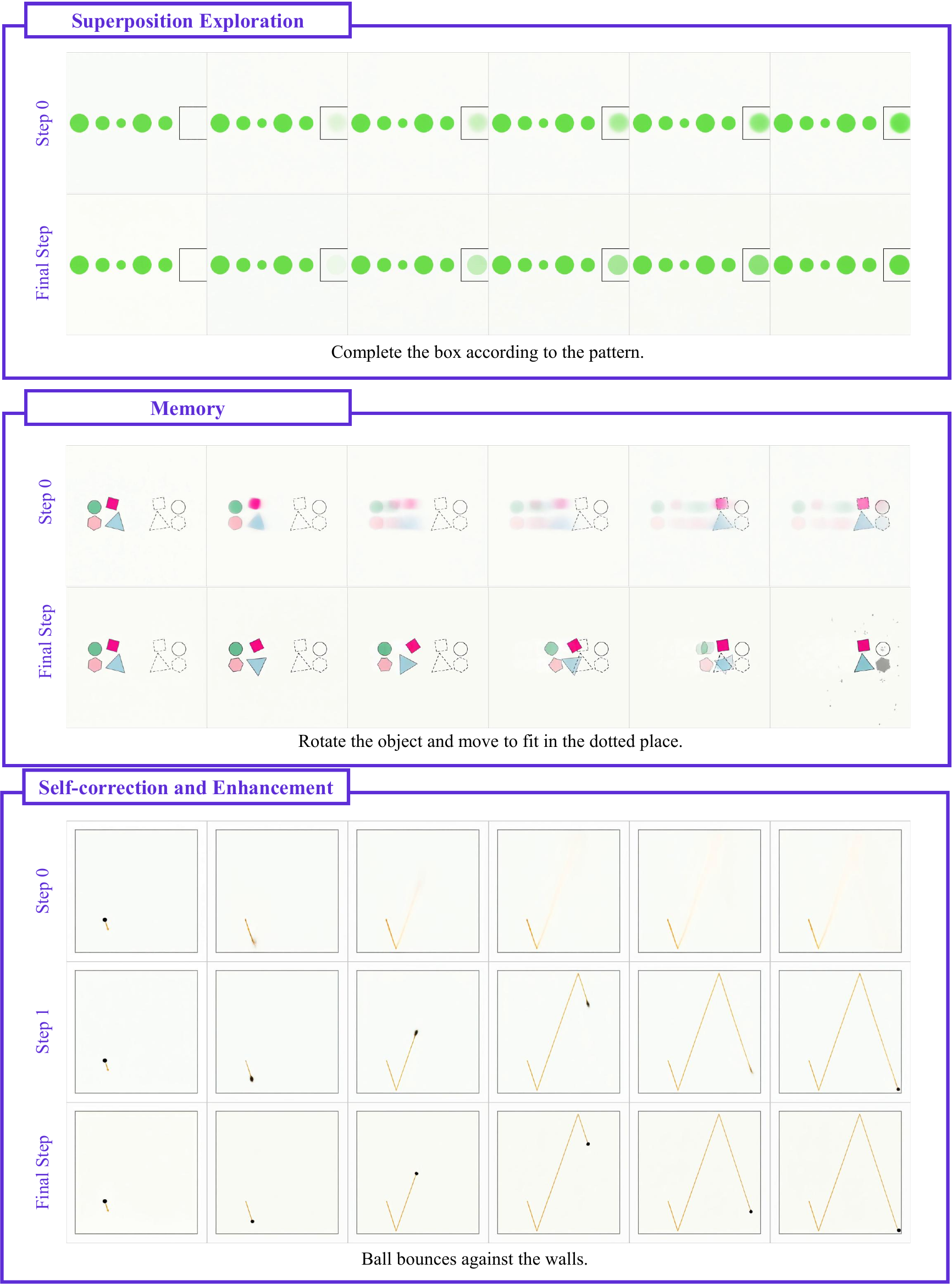}
\caption{Qualitative visualizations of the distilled model.}
\label{fig:distilled_examples}
\end{figure}
These results suggest that even for distilled models, preserving sufficient latent evolution during the initial diffusion step is crucial for maintaining effective reasoning capability.


\section{Qualitative Examples on VBVR-LTX2.3 and VBVR-Wan2.1}
\label{sec:ltx2.3_and_wan2.1}
\cref{fig:viz_ltx,fig:viz_wan21} present video reasoning samples from VBVR-LTX2.3 and VBVR-Wan2.1, respectively. For each model, we show multiple representative examples together with visualizations of the corresponding denoising trajectories.  Consistent with our main findings, both models exhibit clear Chain-of-Steps (CoS) dynamics. That is, reasoning unfolds progressively along denoising steps, intermediate hypotheses emerge and are refined over time, and final conclusions gradually stabilize during later stages of generation. The examples further reveal similar \textit{Working Memory}, \textit{Self-correction and Enhancement} behaviors, suggesting  that the qualitative phenomena identified in the main paper  are broadly shared among diffusion-based video reasoning models.

\section{More Qualitative Examples on VBVR-Wan2.2}
In this section, more visualization examples are provided to further illustrate that reasoning happens across diffusion steps for video generation models. These examples illustrate several recurring phenomena discussed in the main paper.

\begin{figure}[!t]
\vspace*{-0cm}
\centering
\includegraphics[width=0.95\textwidth]{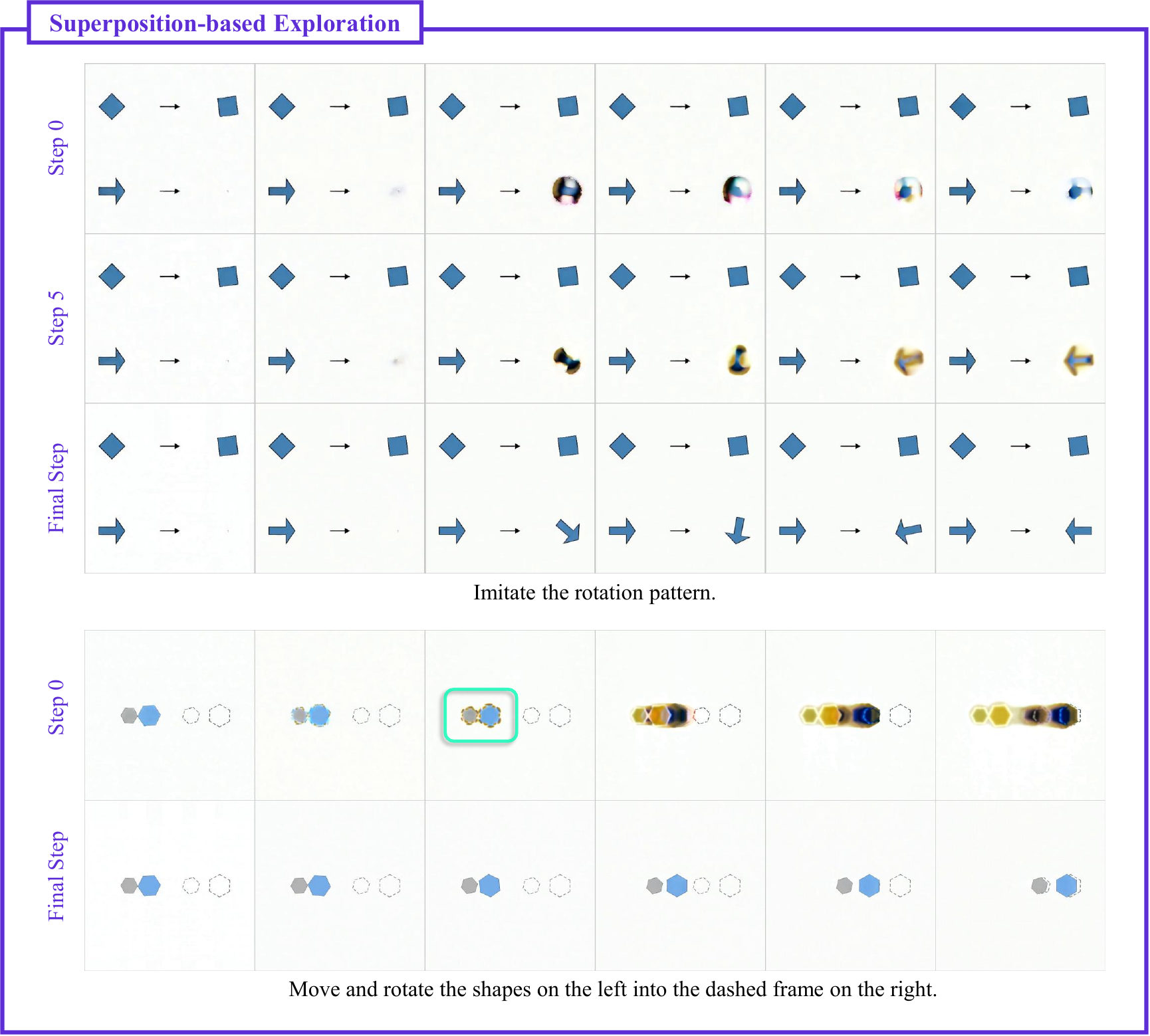}
\caption{More visualizations of "Superposition-based Exploration" phenomenon in \cref{sec:superposition}.}
\label{fig:more_viz_3}
\end{figure}

\begin{figure}[p]
\vspace*{-1.8cm}
\centering
\includegraphics[width=0.95\textwidth]{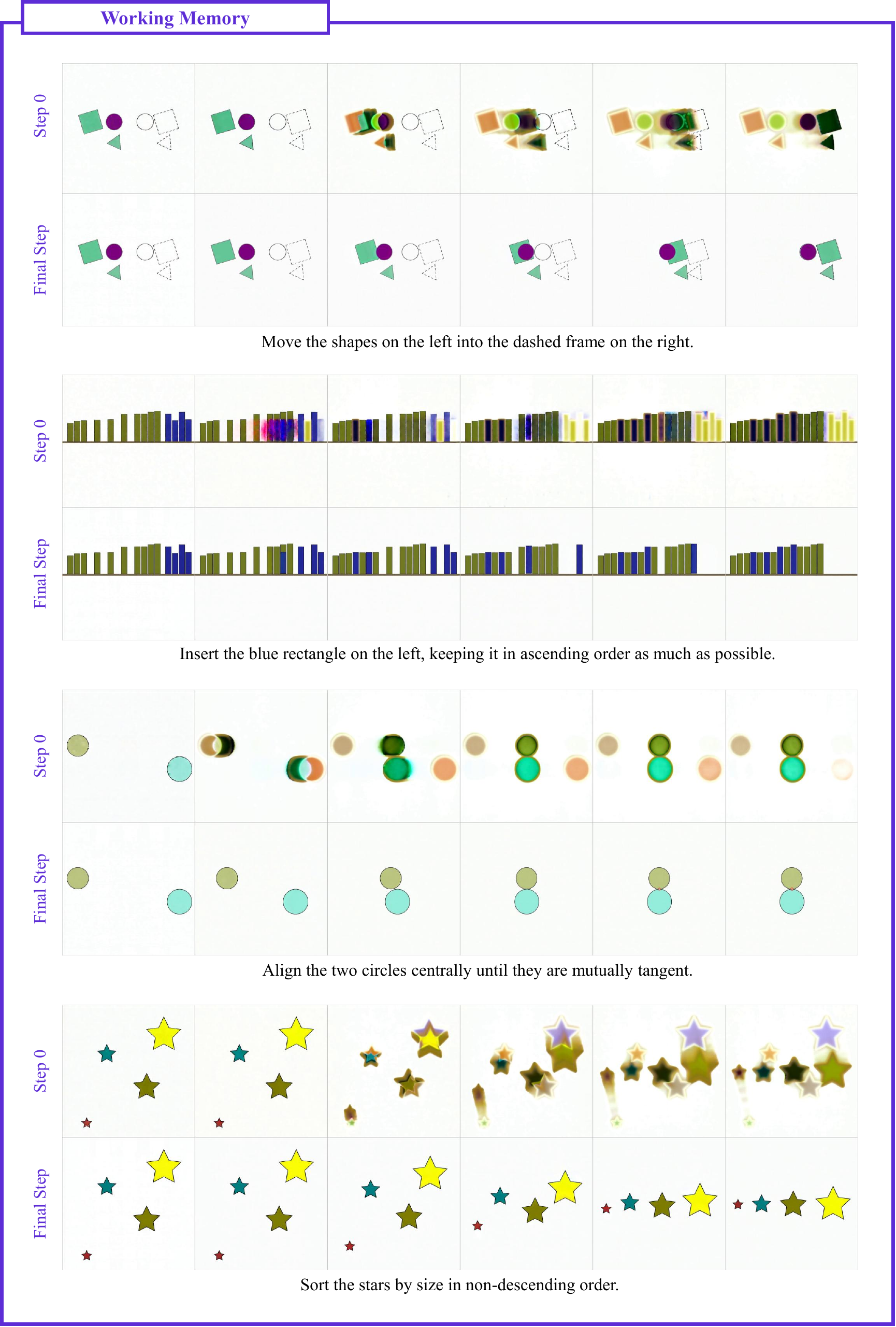}
\caption{More visualizations of "Memory" phenomenon in \cref{sec:memory}.}
\label{fig:more_viz_4}
\end{figure}

\begin{figure}[p]
\vspace*{-1.8cm}
\centering
\includegraphics[width=0.95\textwidth]{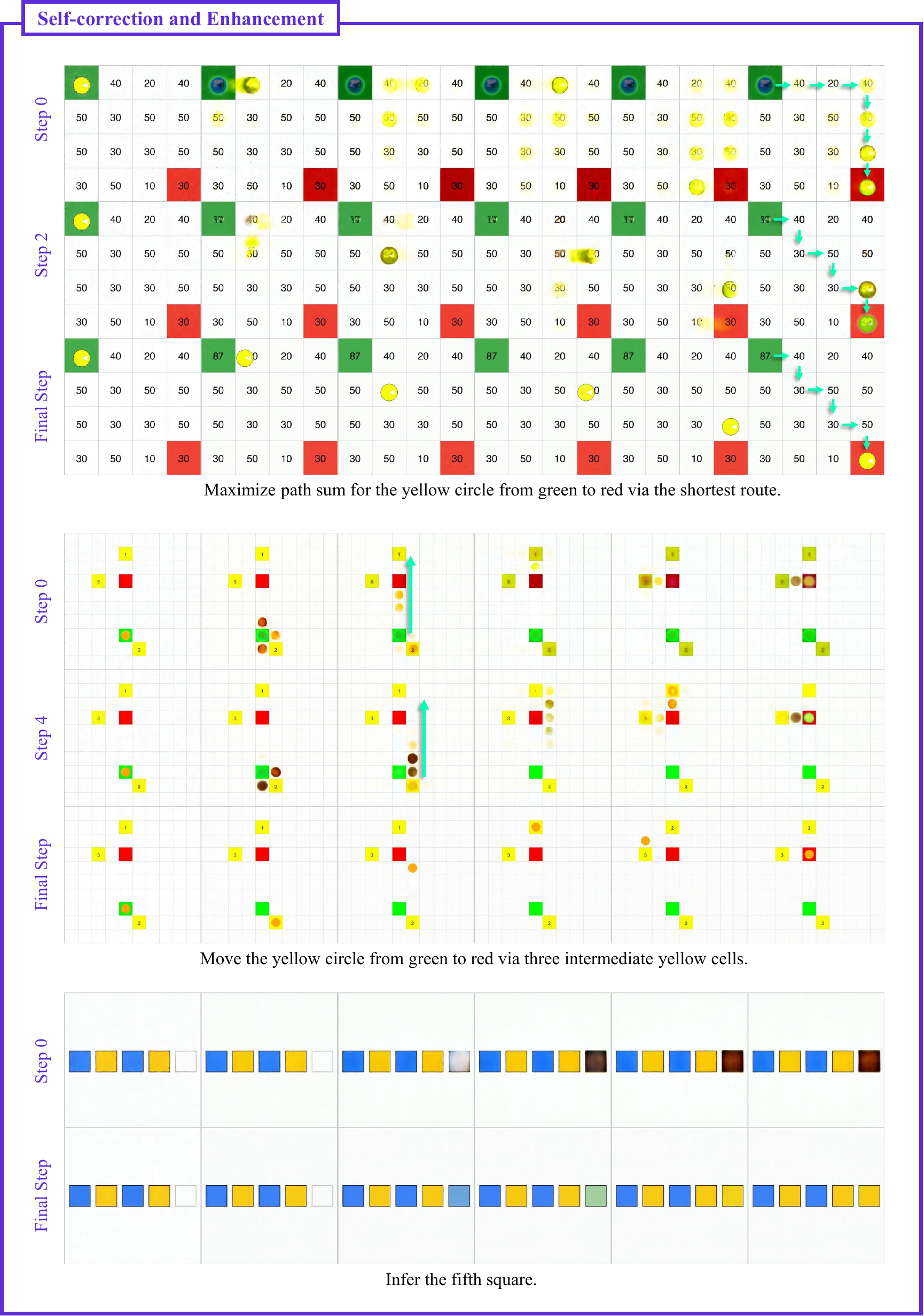}
\caption{More visualizations of "Self-correction and Enhancement" phenomenon in \cref{sec:self-correction}.}
\label{fig:more_viz_5}
\end{figure}

\begin{figure}[t]
\vspace*{-1.8cm}
\centering
\includegraphics[width=0.95\textwidth]{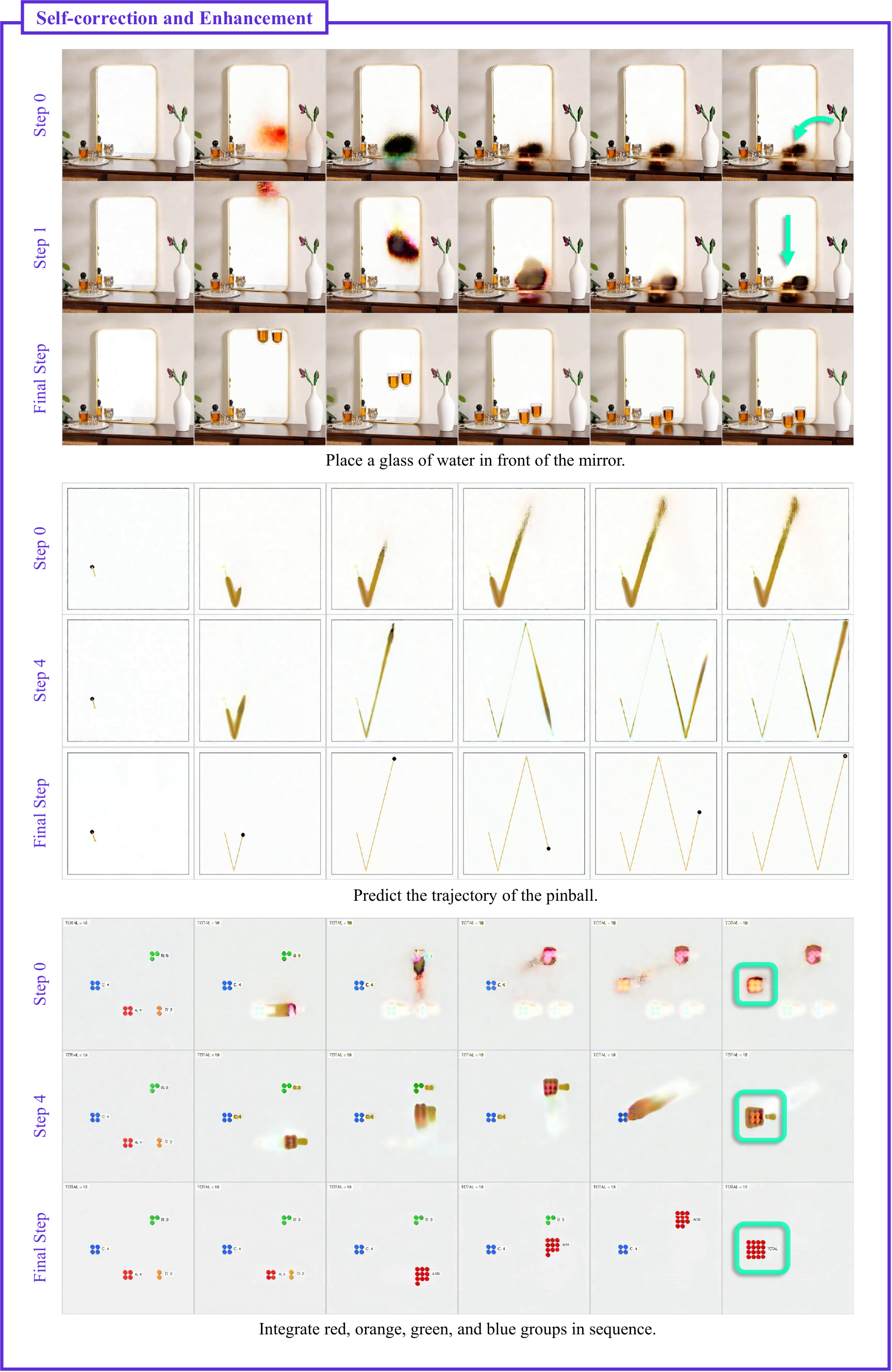}
\caption{More visualizations of "Self-correction and Enhancement" phenomenon in 
\label{fig:more_viz_6}
\cref{sec:self-correction}.}
\end{figure}

Specifically, \cref{fig:more_viz_1,fig:more_viz_2} presents additional cases of the \textit{Multi-Path Exploration} phenomenon described in \cref{sec:multi_path}, where the model explores multiple candidate structures before converging to a coherent generation. \cref{fig:more_viz_3} provides further examples of \textit{Superposition-based Exploration} discussed in \cref{sec:superposition}, highlighting how multiple hypotheses may coexist in intermediate representations. \cref{fig:more_viz_4} illustrates the \textit{Working Memory} phenomenon from \cref{sec:memory}, showing how the model memorises key information such as the trajectory. Finally, \cref{fig:more_viz_5,fig:more_viz_6} demonstrate additional instances of \textit{Self-correction and Enhancement} described in \cref{sec:self-correction}, where early imperfect structures are gradually refined and improved through the denoising steps.
\begin{figure}[ht]
\vspace{-8pt}
\centering
\includegraphics[width=0.95\textwidth]{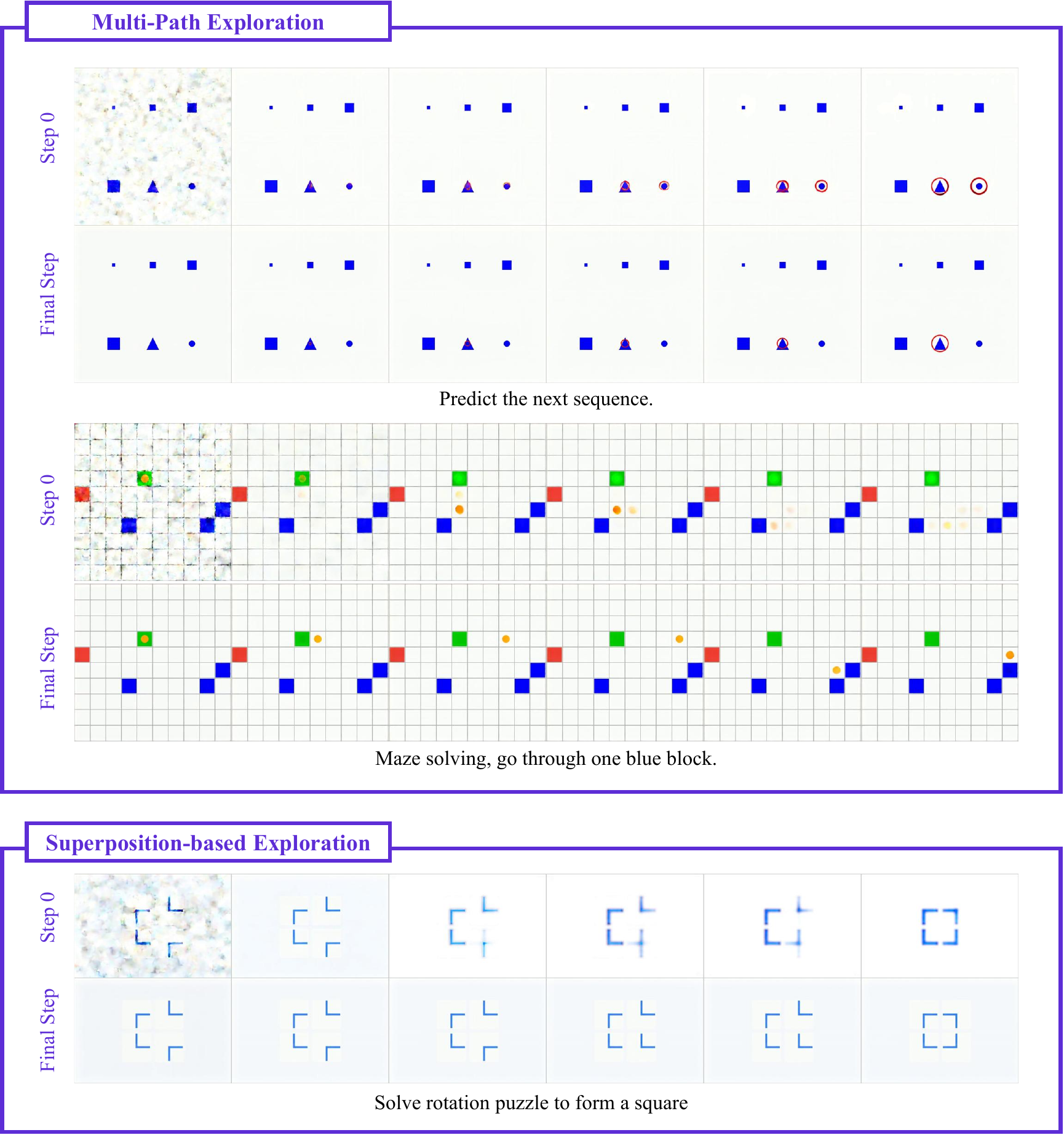}
\caption{Qualitative examples on VBVR-LTX2.3.}
\label{fig:viz_ltx}
\end{figure}

\begin{figure}[ht]\ContinuedFloat
\centering
\includegraphics[width=0.95\textwidth]{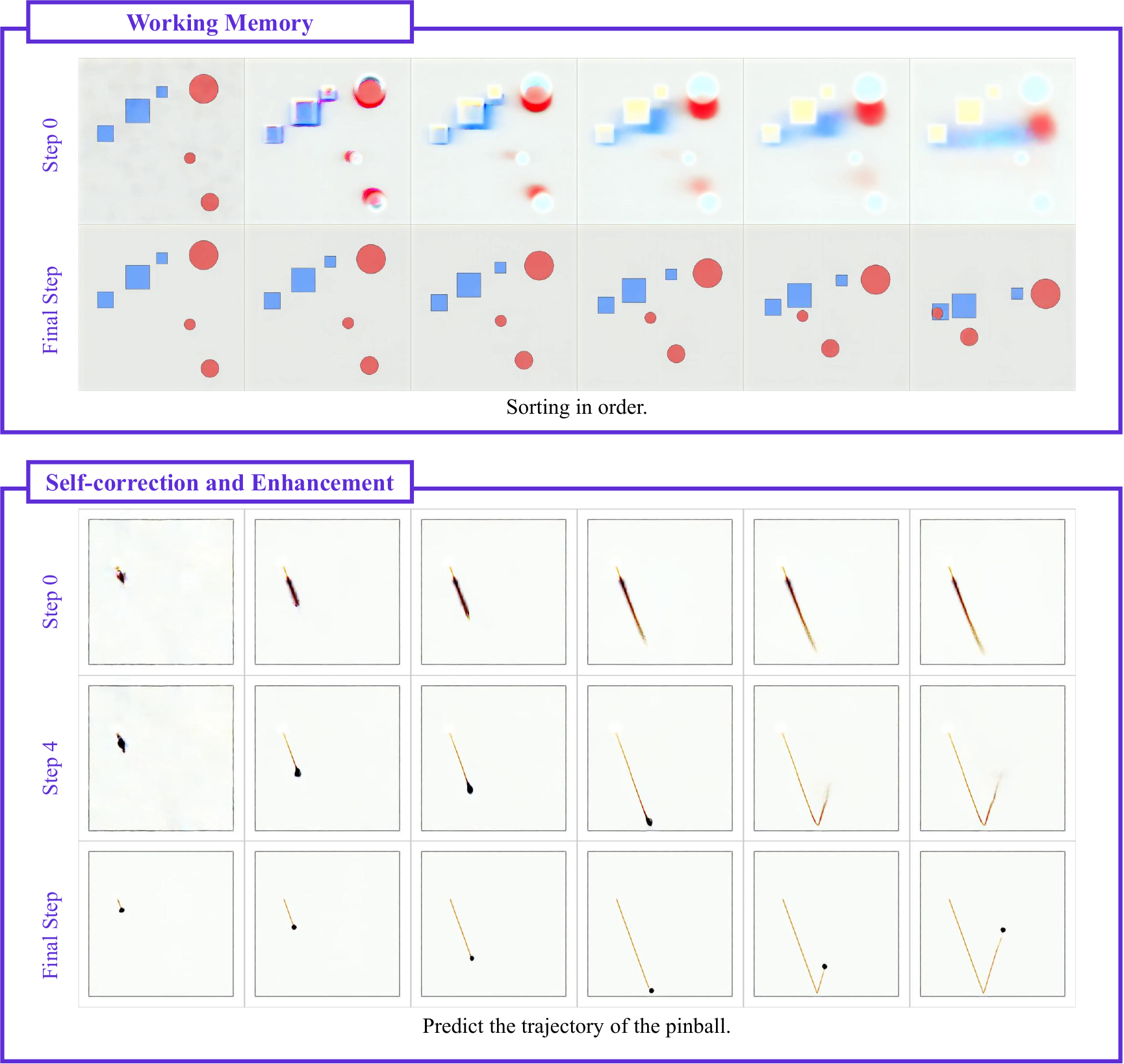}
\caption{(Continued) Qualitative examples on VBVR-LTX2.3.}
\end{figure} 
\begin{figure}[!h]
\vspace{-8pt}
\centering
\includegraphics[width=0.95\textwidth]{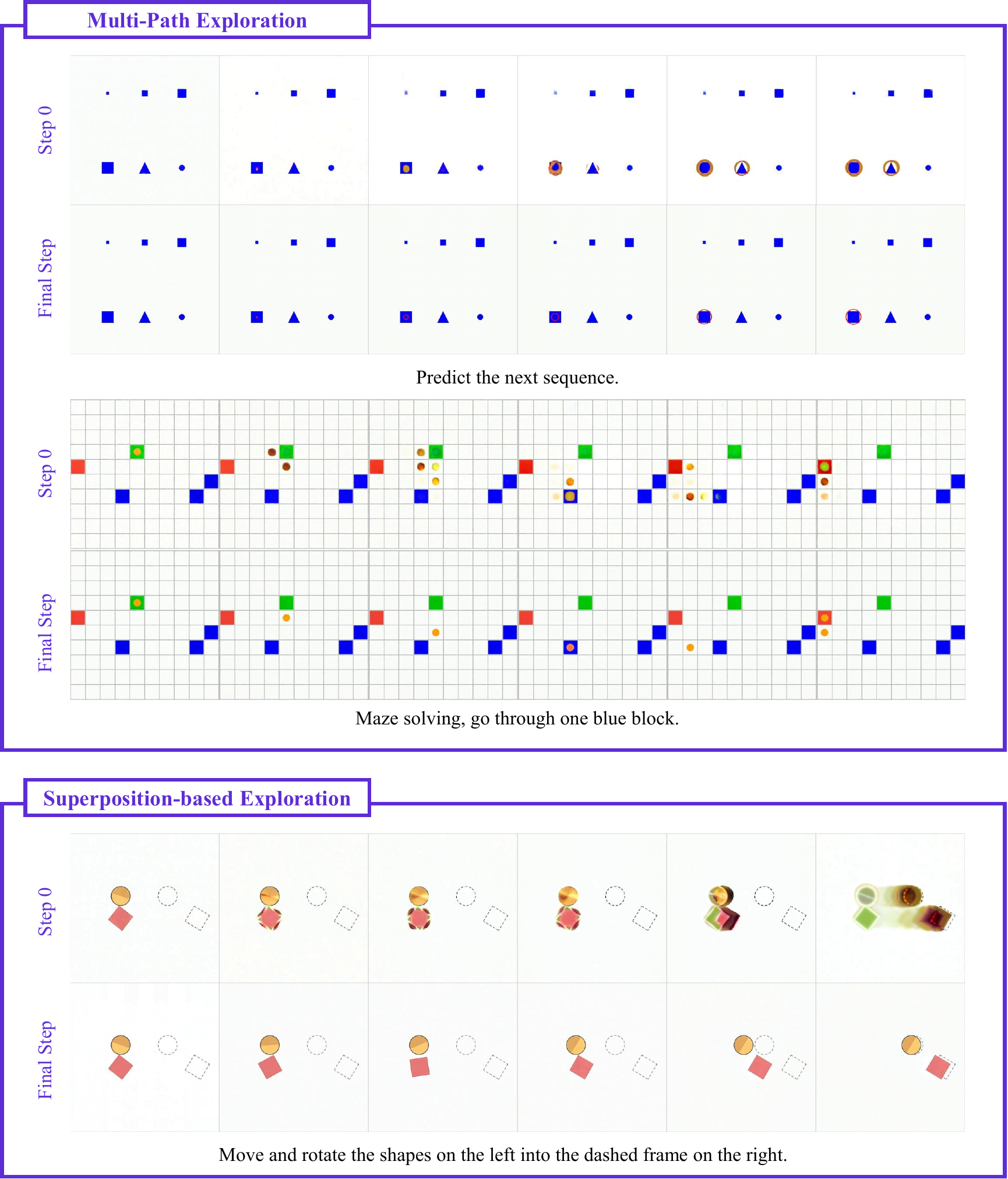}
\caption{Qualitative examples on VBVR-Wan2.1.}
\label{fig:viz_wan21}
\end{figure}

\begin{figure}[!h]\ContinuedFloat
\centering
\includegraphics[width=0.95\textwidth]{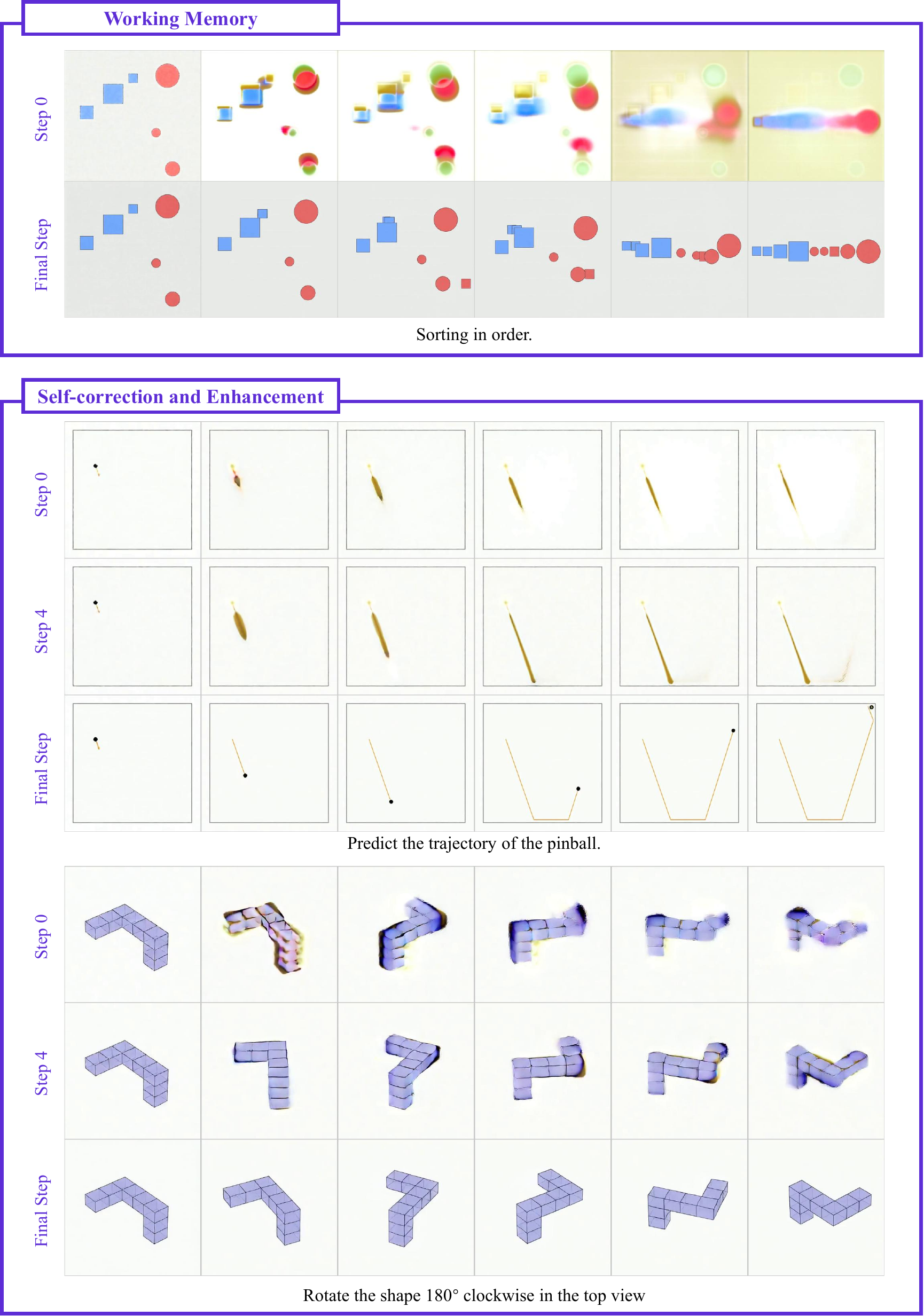}
\caption{(Continued) Qualitative examples on VBVR-Wan2.1.}
\end{figure} 

\begin{figure}[!t]
\vspace*{-3cm}
\centering
\includegraphics[width=0.95\textwidth]{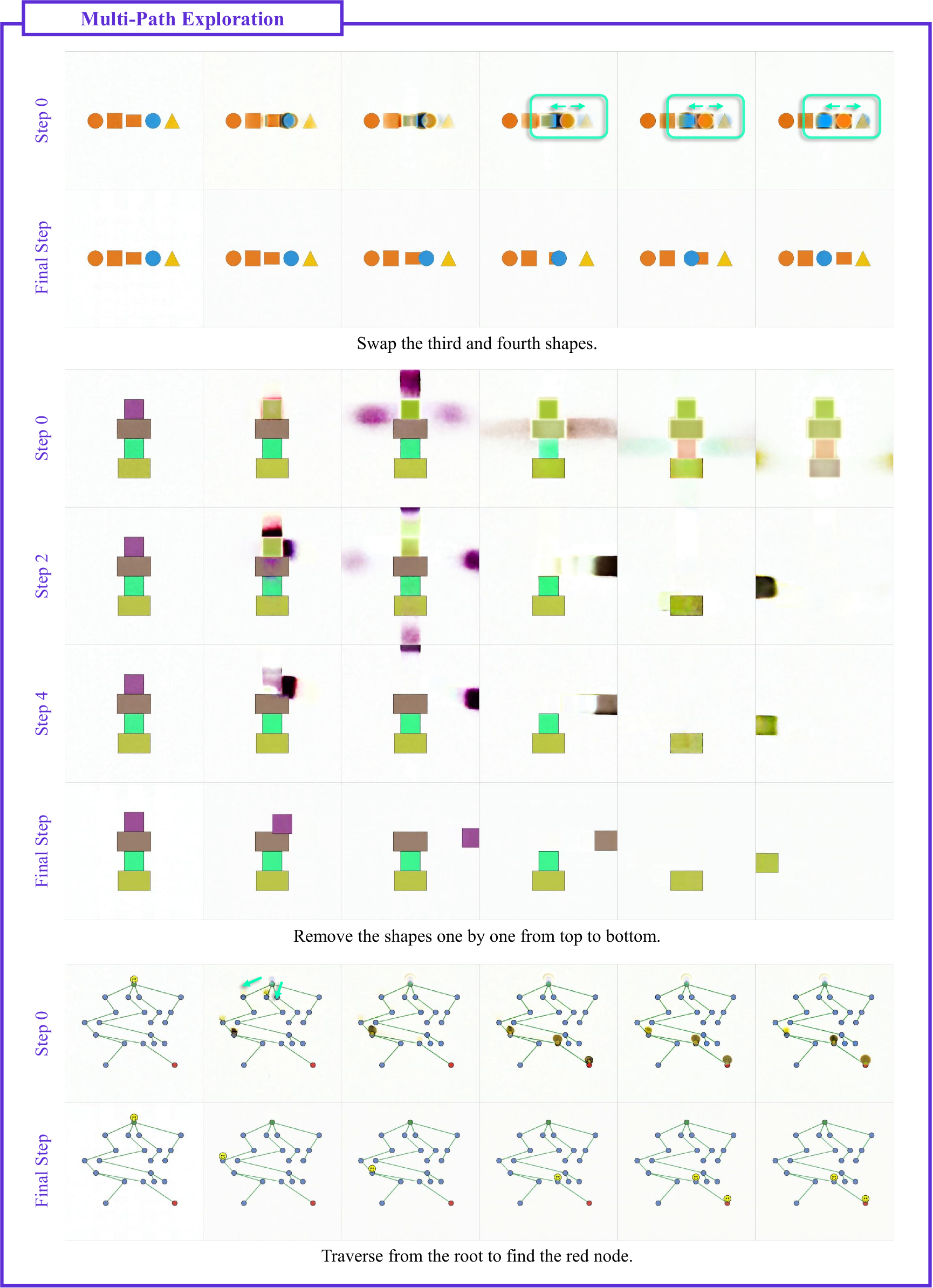}
\caption{More visualizations of "Multi-Path Exploration" phenomenon in \cref{sec:multi_path}.}
\label{fig:more_viz_1}
\end{figure}
\begin{figure}[!t]\ContinuedFloat
\vspace*{-5cm}
\centering
\includegraphics[width=0.95\textwidth]{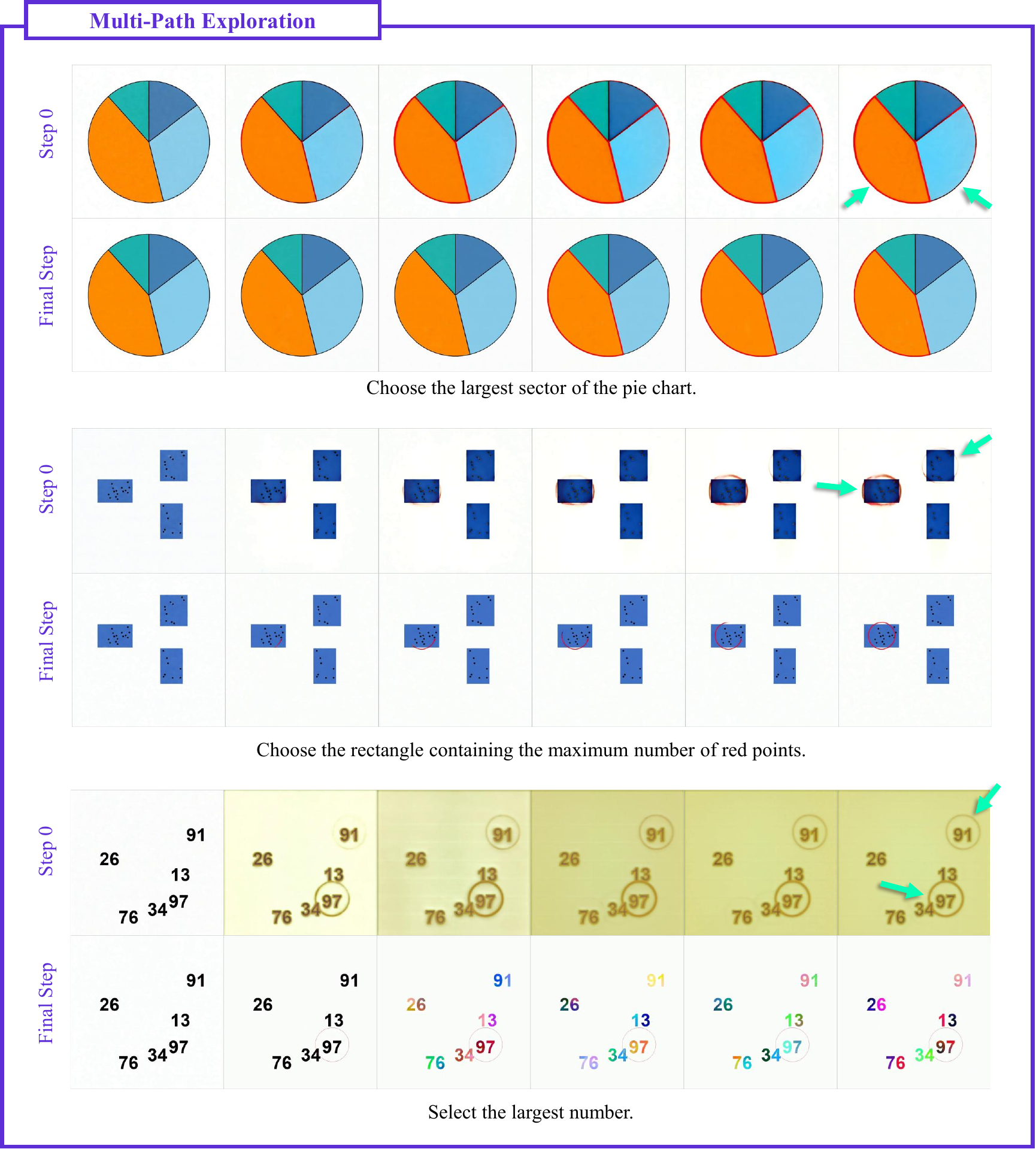}
\caption{(Continued) More visualizations of "Multi-Path Exploration" phenomenon in \cref{sec:multi_path}.}
\label{fig:more_viz_2}
\end{figure}

\end{document}